\documentclass[letterpaper, 10 pt, journal]{IEEEtran}  

\newcommand{\sjw}{\textcolor{black}} 

\usepackage{amsthm}
\usepackage{algorithm, algorithmicx, algpseudocode}
\usepackage{amsmath,amssymb,enumerate}
\usepackage{mathrsfs}
\usepackage{times}
\usepackage{multicol}
\usepackage{amsfonts}
\usepackage{textcomp}
\usepackage{balance}
\usepackage{multirow}
\usepackage{booktabs}
\usepackage{dblfloatfix} 
\usepackage{bbold}
\usepackage{makecell}
\usepackage{hhline} 
\usepackage{float} 
\usepackage[mathscr]{euscript}
\usepackage[sort&compress,numbers]{natbib}

\usepackage{comment}
\usepackage{xparse}
\usepackage{lipsum}
\usepackage{soul} 
\usepackage{balance}
\usepackage{dsfont} 
\usepackage{setspace} 
\usepackage{accents} 

\usepackage{graphicx}
\usepackage{fixltx2e}

\usepackage[T1]{fontenc}
\usepackage[usenames,dvipsnames,svgnames,table, xcdraw]{xcolor}
\usepackage[caption=false,font=footnotesize]{subfig}
\usepackage[utf8]{inputenc}

\usepackage{hyperref}
\hypersetup{
  colorlinks=true,pdfpagemode=UseNone,citecolor=black,linkcolor=black,urlcolor=BrickRed
}

\usepackage{caption}
\graphicspath{
    {fig/}
}

\makeatletter
\newcommand{\mathleft}{\@fleqntrue\@mathmargin0pt}
\newcommand{\mathcenter}{\@fleqnfalse}
\makeatother

\title{BDIS: Bayesian Dense Inverse Searching Method for Real-Time Stereo Surgical Image Matching}%
\author{Jingwei Song, Qiuchen Zhu, Jianyu Lin, and Maani Ghaffari%
	\thanks{J. Song and M. Ghaffari are with the University of Michigan, Ann Arbor, MI 48109, USA. \texttt{\{jingweso,maanigj\}@umich.edu}}%
	\thanks{Q. Zhu is with School of Electrical and Data Engineering, University of Technology, Sydney, P.O. Box 123, Broadway, NSW 2007, Australia. \texttt{Qiuchen.Zhu@uts.edu.au}}%
	\thanks{J. Lin is with Hamlyn Centre for Robotic Surgery, Imperial College London, London SW7 2AZ, UK. \texttt{xjtuljy@gmail.com}}
}

\begin{document}

	\maketitle

	\begin{abstract}
		
		In stereoscope-based Minimally Invasive Surgeries (MIS), dense stereo matching plays an indispensable role in 3D shape recovery, AR, VR, and navigation tasks. Although numerous Deep Neural Network (DNN) approaches are proposed, the conventional prior-free approaches are still popular in the industry because of the lack of open-source annotated data set and the limitation of the task-specific pre-trained DNNs. Among the prior-free stereo matching algorithms, there is no successful real-time algorithm in none GPU environment for MIS. \sjw{This paper proposes the first CPU-level real-time prior-free stereo matching algorithm for general MIS tasks.} We achieve an average $17$ Hz on $640 \times 480$ images with a single-core CPU (i5-9400) for surgical images. Meanwhile, it achieves slightly better accuracy than the popular ELAS. The patch-based fast disparity searching algorithm is adopted for the rectified stereo images. A coarse-to-fine Bayesian probability and a spatial Gaussian mixed model were proposed to evaluate the patch probability at different scales. An optional probability density function estimation algorithm was adopted to quantify the prediction variance.   Extensive experiments demonstrated the proposed method's capability to handle ambiguities introduced by the textureless surfaces and the photometric inconsistency from the non-Lambertian reflectance and dark illumination. The estimated probability managed to balance the confidences of the patches for stereo images at different scales. It has similar or higher accuracy and fewer outliers than the baseline ELAS in MIS, while it is 4-5 times faster. The code and the synthetic data sets are available at \url{https://github.com/JingweiSong/BDIS-v2}.
		
		
	\end{abstract}

	\section{Introduction}
	
	\IEEEPARstart{M}{inimally} Invasive Surgery (MIS) technique is widely adopted in modern surgery since it mitigates postoperative infections and enables faster recovery of the patients. However, the surgeons suffer from the small field of view because the procedures are performed in a narrow space with elongated tools and without direct 3D vision. Hence, MIS poses more difficulties to surgeons than open surgeries~\cite{mountney2009dynamic}. To overcome this challenge, stereoscopes are integrated into the operational imaging system to provide 3D stereo instead of single 2D images. Moreover, the recovered 3D shape can be further applied in applications including dense Simultaneous Localization and Mapping (SLAM)~\cite{song2017dynamic,song2018mis}, AR~\cite{haouchine2013image,widya20193d} and diseases diagnosis~\cite{mahmood2019polyp,jia2020automatic}. \sjw{Among the many off-the-shelf stereo matching methods, real-time implementation \cite{edwards2021challenge} is essential for tasks like surgeon-centered AR, reduction of error, decision making, or autonomous surgery's safety boundary. We would like to emphasize that the term ``real-time'' is task-dependent and has no clear definition of frame rate in the robotic community. This research defines ``real-time'' as over 10 Hz, which is enough to serve most computer-aided tasks in MIS or robotic surgery, such as visualization and surgical navigation..}\par
	
	\sjw{The state-of-the-art stereo intra-operative tissue shape reconstruction techniques strictly follow the pin-hole camera projection model~\cite{andrew2001multiple} and bridge the 3D shape and 2D stereo image. The major difference exists in the workflow for estimating the disparities from the left-right image pairs.} Stereo matching algorithms can be classified as prior-free and prior-based. Prior-free approaches refer to aligning left and right images pixel-wisely using explicit handcrafted features for corner points matching (feature-metric), photometric consistency presumption for direct dense pixel searching (photo-metric)~\cite{kroeger2016fast}, or the combination of both methods~\cite{hirschmuller2005accurate,stoyanov2010real,geiger2010efficient,rappel2016surgical}. Differently, incorporated with the annotated disparity, prior-based methods, mostly Deep Neural Network (DNN) based approaches, directly learn the complex ``images to disparity'' process with the black box model from the training data set~\cite{ye2017self,turan2018deepvo,yang2019hierarchical,tonioni2019real,xu2020aanet,brandao2020hapnet,long2021dssr}.\par 
	
	The prior-free and prior-based categories do not contradict. Although DNN-based methods~\cite{ye2017self,turan2018deepvo} are reported to be more efficient, it comes with several disadvantages. First, predictions from DNN-based methods may be invalidated with changing parameters such as focal length and baseline or a significant difference in the texture between the training and testing data~\cite{pratt2014practical,allan2021stereo}. \sjw{Moreover, the computational resources are insufficient for DNN-based methods when lacking high-end GPU or saving GPU for other tasks such as SLAM~\cite{song2018mis}, detection, segmentation, or disease diagnosis~\cite{jia2020automatic}. Computational resources deficiency in the operating room is a common problem for medical devices in the operating room. Integrating more computational devices is difficult due to hardware design constraints in thermal and physical dimensions. Thus,} the lack of a powerful GPU severely limits the performance of DNN in practice. More importantly, in many cases, access to high-quality training data is unavailable because of lacking the necessary hardware or other ethnic issues. This drawback could even lead to a failure in training and predicting. Therefore, prior-free methods are still widely applied in the industry for their robustness and are free of the labeled training data. \par

	Among the prior-free methods, ELAS~\cite{geiger2010efficient} is the most widely used prior-free algorithm in MIS. Both industry~\cite{zampokas2018real,cartucho2020visionblender} and academy~\cite{song2017dynamic,song2018mis,zhang2017autonomous,zhan2020autonomous} routinely adopt ELAS for depth recovery due to its satisfying accuracy and robustness. However,  CPU-based ELAS can only process stereo matching in near real-time. ELAS is adopted to serve real-time stereo requirements on GPU, which is not always available in the operating rooms. \sjw{Therefore, this article aims at developing a real-time (more than 10 Hz on $640 \times 480$ on modern CPU) and more accurate general stereo matching algorithm for CPU-level real-time stereo matching, which can substitute ELAS.} \par

	The work Dense Inverse Searching (DIS)~\cite{kroeger2016fast}, which was proposed to conduct real-time optical flow, provided valuable insights on fast disparity searching. DIS demonstrated that the Lucas-Kanade (LK) optical flow algorithm~\cite{lucas1981iterative} could be modified to a fast disparity searching manner. Meanwhile, it also showed that the accuracy could be improved significantly by implementing the fast LK using the coarse-to-fine patch registration. Therefore, coarse-to-fine fast LK has the potential to be applied for estimating the disparities from images recording the continuous surfaces, such as the MIS scenario. However, the basic assumption of fast LK, the left-right image photometric consistency, cannot be fully satisfied in MIS. \textbf{Three factors contribute jointly to the inconsistency: textureless surface, dark region, and non-Lambertian reflectance.} Unlike the indoor/outdoor scenario with abundant texture,  surgical scenes contain textureless soft-tissue surfaces. Furthermore, the point source illumination of the scope leads to some surface regions in the dark, which in turn exacerbates the textureless issue. \sjw{The non-Lambertian reflectance brings uneven brightness to the surfaces, and it cannot be eliminated by just enforcing the patch normalization~\cite{shimasaki2013generating}.} Previous researches~\cite{engel2015large,engel2017direct} demonstrated that the non-Lambertian reflectance could be modeled with an affine lighting model on the observed images. Consequently, the accuracy of the navigation system can be notably improved in the indoor scenario. Nevertheless, the affine modeling strategy cannot be directly applied to Computer Assisted Surgery (CAS). Unlike the indoor/outdoor cases, \citet{shimasaki2013generating} pointed out that the photometric consistency in MIS is severely contaminated, and the relationship is much more complicated than the affine modeling.\par

	This article extends our preliminary work~\cite{song2021bayesian} and proposes Bayesian Dense Inverse Searching (BDIS), the first prior-free CPU-level real-time stereo matching algorithm, to achieve comparable accuracy as the state-of-the-art near real-time ELAS algorithm. The deterministic fast LK algorithm from~\cite{kroeger2016fast} is adopted and deeply integrated with our patch-wise Bayesian posterior probability, which is based on the Conditional Random Fields (CRFs). \sjw{Moreover, a spatial simplified Gaussian Mixture Model (sGMM) is adopted to quantify pixel-wise confidence within the patch.} A probability propagation strategy is designed to allow the probability to cover all scales. \sjw{The probabilities quantify the confidence of multiple overlapping disparities of DIS patch suffering from the textureless surface, dark region, and non-Lambertian reflectance in MIS. To overcome the 3 major factors mentioned above, the probability measuring module is employed for both local overlapping patch estimation fusion and outlier filtering. In general, this work has the following contributions: }\par
	
	\begin{itemize}
		\item To our knowledge, BDIS is the first single-core CPU stereo matching approach that achieves similar performance to the near real-time method ELAS.
		\item A computationally-economic probabilistic model is developed to quantify the posterior probability of the patch based on a simplified Bayesian assumption.  
		\item \sjw{An uncertainty-aware estimation of the disparity using spatial sGMM is proposed to quantify the pixels' confidence within the patch.} 
		\item A coarse-to-fine probability propagation algorithm is proposed to quantify the patch from different scales' perspectives.
		\item \sjw{A Maximum A Posteriori (MAP) based Probabilistic Density Function (PDF) estimation algorithm is explored, which can correctly estimate the pixel-wise variance if the assumption of Gaussian distribution is valid and fastLK converges close to global minima.}
		\item An open-source C++ implementation is released	along with the synthesized stereo images and depths.
	\end{itemize}
	
	This article is an extension of our preliminary work~\cite{song2021bayesian} and contributes additionally in the following ways. Unlike the separate probability estimation in each level, this article proposes a coarse-to-fine strategy to allow the finer prediction to encode the probability of its coarser parents. Moreover, a dynamic variance selection strategy is proposed for CRFs-based probability estimation. Based on the probability within the window, a MAP-based PDF estimation is proposed to quantify the pixel-wise variance. \sjw{Finally, an extensive amount of in-vivo/ex-vivo real-world experiments with reference depth \footnote{\sjw{In the robotic community, ``true value'' or ``reference value'' obtained from hardware/software with higher precision is often termed as ``ground truth''. Following \cite{EDWARDS2022102302}, we name it as ``reference'' considering its precision inaccuracies. The rest content, figures, and tables use ``reference'' instead of ``ground truth''.}} and several ablation studies are conducted to validate the performance of the proposed method.}\par

	The remainder of this article is organized as follows. Section~\ref{section_relatedworks} provides an overview of the related work. Section~\ref{secion_methology} covers the methodology with all the technical details. \sjw{Section~\ref{section_result} conducts experiments to validate the proposed method thoroughly. These include the qualitative and quantitative tests on the synthetic, in-vivo data set, and ex-vivo data set. An ablation study is conducted to illustrate the contributions of different modules.} Lastly, Section~\ref{section_conclusion} concludes this article. \par

	\section{Related works}
	\label{section_relatedworks}
	Stereo matching is one of the most heavily investigated topics in computer vision~\cite{scharstein2002taxonomy}. Trying to mimic human vision, the stereo system comprises two cameras and a computing device to triangulate the 3D shape by estimating the parallax between the left and right images. Efficient and accurate stereo matching algorithms are essential or helpful for many tasks in MIS, such as scope navigation, AR/VR, and disease diagnosis. Regarding the theory and requirement, stereo matching algorithms can be categorized into two groups, i.e., prior-free and prior-based methods.\par
	
	\textbf{Prior-free methods.} The traditional prior-free methods directly estimate the parallax by addressing the similarity between the left and right image. The similarity refers to the hand-crafted features, illumination invariance, or other specially designed metrics such as zero-mean normalized cross correlation~\cite{lin2017optimizing}. \sjw{The global methods~\cite{sun2003stereo,hirschmuller2005accurate,klaus2006segment,hirschmuller2007stereo} minimize a cost function that contains a similarity data term and a smoothness term.} The similarity data term enforces the illumination invariance or some similarity metrics, while the smoothness term regularizes the disparities between neighboring pixels. To save the computational resources, local approaches~\cite{yoon2006adaptive,min2011revisit,zhang2009cross,yang2012non} simplify the optimization by aggregating neighboring matching costs of the pixels. Among these approaches, the Semi-Global Block Matching (SGBM)~\cite{hirschmuller2005accurate,hirschmuller2007stereo} is one of the most widely applied methods in the academy and industry. SGBM constructs the matching cost volume with the range of the predefined depth. Then, it aggregates the cost with the winner-takes-all strategy and computes the disparity directly without iterative optimizations in the global methods. \sjw{The recovered depth is further refined with various strategies, such as eliminating small patches and filtering low contrast pixels.} The aggregation and winner-takes-all strategies ensure SGBM obtains high-quality depth without heavy computational burdens in the optimization step. ELAS~\cite{geiger2010efficient}, which deviates from these cost-volume based depth searching techniques, has also been widely used in industry~\cite{zampokas2018real,cartucho2020visionblender} and academy~\cite{song2017dynamic,song2018mis,zhang2017autonomous,zhan2020autonomous}. Its procedure consists of sparse and dense steps. In the sparse matching, ELAS uses Sobel masks to conduct sparse corner matching as the supporting points set. The aligned sparse supporting points are used for Delaunay triangulation to initialize pixel-wise disparity. After the initialization, a densification step is carried out by maximizing the posterior probability defined by the photometric consistency. ELAS requires around $0.25-1$ second with a single-core modern CPU core. \sjw{Thus, the real-time version needs to be implemented on the GPU end (the code on GPU is publicly available)}. Aiming at real-time disparity estimation, DIS~\cite{kroeger2016fast} proposed a fast LK searching technique for stereo matching. The fast LK, combined with the coarse-to-fine strategy, estimates the parallax in real-time. Nevertheless, the deterministic fast LK was not applied in the MIS community because it is vulnerable to imperfect image pairs contaminated by textureless surface, non-Lambertian reflectance, and dark illumination. Thus, this research adopts the deterministic fast LK algorithm and proposes a probabilistic formulation that is robust to these obstacles.\par

 	\textbf{Prior-based methods.} With the presence of high-quality training data set, prior-based end-to-end stereo matching methods also serve the stereo matching tasks. These DNN-based methods have an advantage over the traditional prior-based methods in their ability to learn more complex end-to-end searching. The delicately designed hand-crafted similarity term in traditional prior-based methods cannot fully handle the complex image to parallax process. The DNN-based methods, however, directly learn the non-linear relationships from the annotated training data set. The first widely used DNN-based method GCNet~\cite{kendall2017end} follows the conventional stereo matching algorithm such as SGBM by building a 3D cost volume based on the left and right feature map. The disparity is obtained by searching the cost volume. PSMNet~\cite{chang2018pyramid} further improves GCNet by introducing pyramid spatial pooling and more convolution layers for cost aggregation. It is reported to have better accuracy over GCNet. Later, GwcNet~\cite{guo2019group} modifies the structure of the 3D hourglass and introduces group-wise correlation to form a group-based 3D cost volume. In CAS domain,~\cite{ye2017self} is one of the earliest researches which adopt DNN-based stereo matching techniques in the CAS domain. PSMNet~\cite{chang2018pyramid} and GwcNet~\cite{guo2019group} also attract the attention from the CAS community~\cite{allan2021stereo}. Other stereo matching techniques~\cite{yang2019hierarchical} and~\cite{brandao2020hapnet} were also tested and recommended by the CAS community, which are capable of accomplishing the stereo matching tasks. The latest work~\cite{long2021dssr} proposed a transformer-based stereoscopic depth perception algorithm.\par
	
	The research of the prior-free CPU-level real-time stereo matching techniques still significantly benefits the MIS community. Despite their efficiency, the DNN-based methods require a decent amount of annotated high-quality data set for training. \sjw{First, the data set is extremely difficult to obtain in the medical domain due to ethnic and hardware limitations. For example, the publicly available data set with the reference~\cite{allan2021stereo} adopted a structured light alone with the endoscope to collect the image and shape from the porcine.} The image and depth are aligned based on the kinematic information of both sensors. Moreover, to solve the synchronization problem of the two sensors, the data is collected in the porcine, which remains still during the entire procedure. Their complicated data collection procedure in porcine implies the greater difficulty in conducting the same process in the human body. Second, the limited (high-end) GPU resource also limits the real-time prediction of these approaches. Take PSMNet as an example, it costs about 4G memory and around 400ms to predict a KITTI stereo pair even on high-end GPUs. Besides, other tasks also involve the computation on GPU, which makes it more difficult to maintain the GPU requirement of DNN-based methods. Therefore, the prior-based and prior-free methods complement each other in different scenarios (with or without qualified training data set). Thus, we aim at proposing the first CPU-level real-time stereo shape recovery algorithm for the MIS community. \par.

	\begin{figure*}[!h]
		\centering
		\subfloat{
			\begin{minipage}[]{1\textwidth}
				\centering
				\includegraphics[width=1\linewidth]{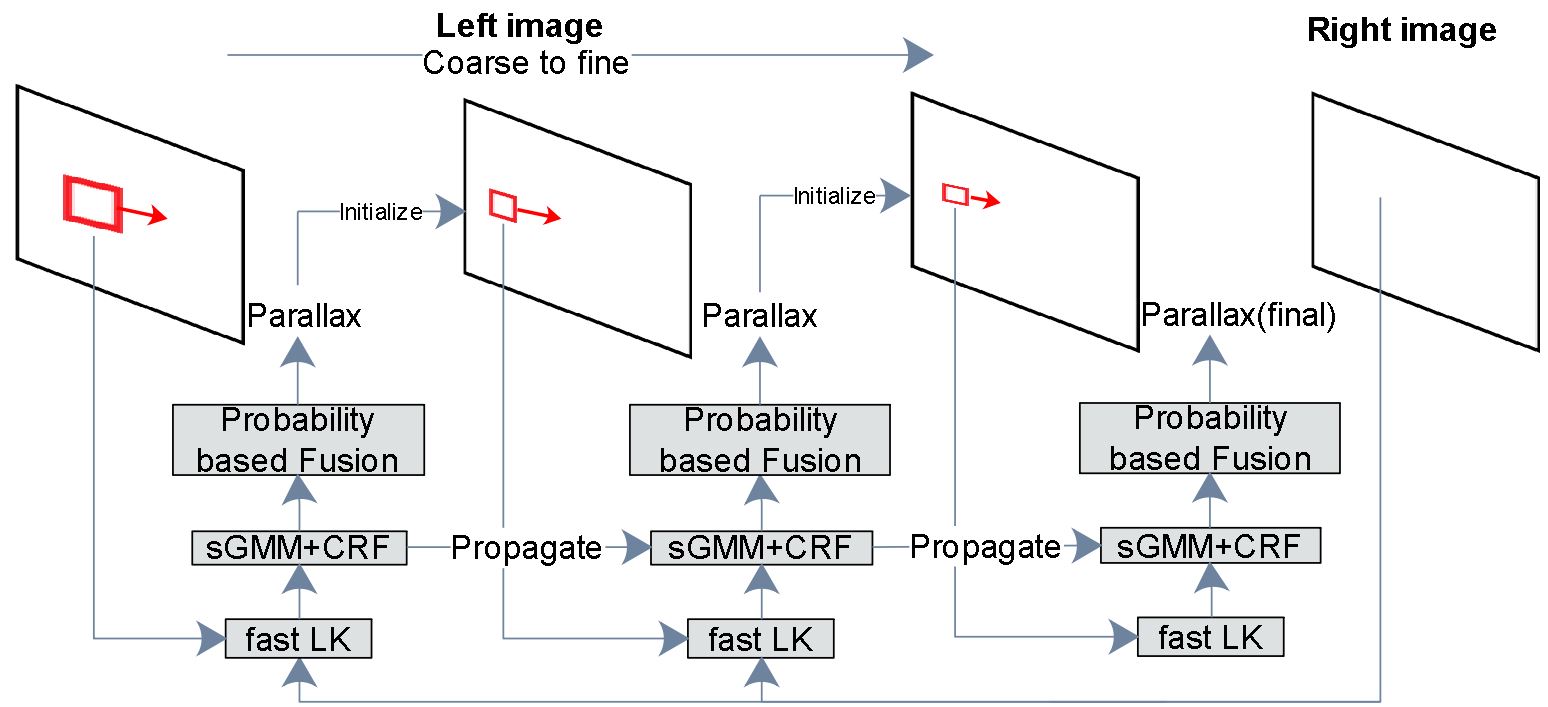}
			\end{minipage}
		}
		\caption{\sjw{Illustrated is the framework of the proposed BDIS. It uses three scale levels as an example. The parallax of the patch is initialized with the prediction from the last level. The fast LK algorithm is applied to estimate the disparity. Then, our sGMM and CRFs algorithms are used to yield the probability of the patch in the current scale as well as the probability propagated from the last level. Finally, the predictions from the batches are fused by addressing the predicted probabilities.}}
		\label{fig_framework}
	\end{figure*}

	\section{Methodology}
	\label{secion_methology}
	
	Fig.~\ref{fig_framework} demonstrates the workflow of the proposed BDIS. The left to right parallax estimation starts from the coarsest level. A fast LK algorithm is applied to estimate the initial disparity of the patch. \sjw{Then, our sGMM and CRFs-based probability propagation module evaluate the appropriate pixel-wise probabilities based on the predictions from the fast LK.} Next, the predictions of all overlapping patches are fused to yield the optimal probability, which is then used as an initialization for the next finer-scale processing. The fusion module strictly addresses the pixel-wise probability. The matching and fusion process is iteratively implemented until it reaches the finest level. In the probability estimation procedure, the probability in the coarser level is propagated to the finer level to account for multi-scale confidences.

	\subsection{The fast LK algorithm}
	\label{section_2_1}
	
	The fast LK was proposed by~\cite{baker2004lucas,kroeger2016fast}. For completeness, we briefly introduce the fast LK algorithm. In the authentic LK~\cite{lucas1981iterative}, the optimal parallax searching is realized by updating the parallax $\mathbf{u}_k^{(n)}$ ($k$th patch on scale level $(n)$) by iteratively searching the optimal parallax $\Delta \mathbf{u}^{\prime}$. This is achieved by minimizing the following objective function, which is

	\begin{equation}
	\label{Eq_fast_inverse_search}
	\Delta \mathbf{u}^{\prime}=\operatorname{argmin}_{\Delta \mathbf{u}^{\prime}} \sum_{x}\left[I^{(n)}_{r}\left(\mathbf{x}+\mathbf{u}+\Delta \mathbf{u}^{\prime}\right)-I^{(n)}_{l}(\mathbf{x})\right]^{2},
	\end{equation}
	
	\noindent where $\mathbf{x}$ is the processed center position of the left patch, $\mathbf{u}$ is the estimated parallax at one searching stage, and it keeps being updated in each loop, $I^{(n)}_l$ and $I^{(n)}_r$ are the left image patch and entire right image. \par
	
	Since the discrete image introduces nonlinear behavior to the system~\eqref{Eq_fast_inverse_search}, the linearization of~\eqref{Eq_fast_inverse_search} is implemented on $\Delta \mathbf{u}^{\prime}$ regarding $I_{r}^{(n)}(\cdot)$ at position $\mathbf{u}$. Therefore, $\mathbf{u}$ is updated in each iteration, and the Jacobian and Hessian matrix need to be re-evaluated in each iteration corespondingly.~\citet{baker2004lucas,kroeger2016fast} proposed the inverse of the roles between left image patch $I_l^{(n)}$ and right image $I_r^{(n)}$. That is switching the roles and optimizing $\sum_{x}\left[I_{r}^{(n)}\left(\mathbf{x}+\mathbf{u}\right)-I_{l}^{(n)}(\mathbf{x}+\Delta \mathbf{u}^{\prime})\right]^{2}$. In the new formulation, the linearization is carried out on the left image patch $I_{l}$, which can be predefined. Therefore, the linearization of the left image patch can be predetermined, and the Jacobian and Hessian only need to be calculated once in the entire optimization process.\par 
	
	After the patch-wise disparity searching, the predictions of the overlapping patches are fused. In DIS~\cite{kroeger2016fast}, the optimal disparity at the location $\mathbf{x}$ was fused by the overlapping patch predictions based on the deterministic weights. The fusion weights were determined by the normalized inverse of the left-to-right photometric residuals. The optimal disparity is

	\begin{equation}
	\label{Eq_residual_fusion}
	\hat{\mathbf{u}}_\mathbf{x}^{(n)} = \sum_{k \in \Omega} \frac{ 1/\lVert I^{(n)}_l(\mathbf{x}+\mathbf{u}_k^{(n)})-I^{(n)}_r(\mathbf{x})\lVert^2}{\sum_{k \in \Omega} 1/\lVert I^{(n)}_l(\mathbf{x}+\mathbf{u}_k^{(n)})-I^{(n)}_r(\mathbf{x})\lVert^2}  \mathbf{u}_k^{(n)},
	\end{equation}
	\noindent where $\Omega$ is the set of patches covering the position $\mathbf{x}$. The pixel-wise disparity $\hat{\mathbf{u}}_\mathbf{x}^{(n)}$ is the weighted average of the estimated disparities from all patches, wherein the weight is the inverse residual of brightness. 
	\begin{figure}[!h]
		\centering
		\subfloat{
			\begin{minipage}[]{0.5\textwidth}
				\centering
				\includegraphics[width=1\linewidth]{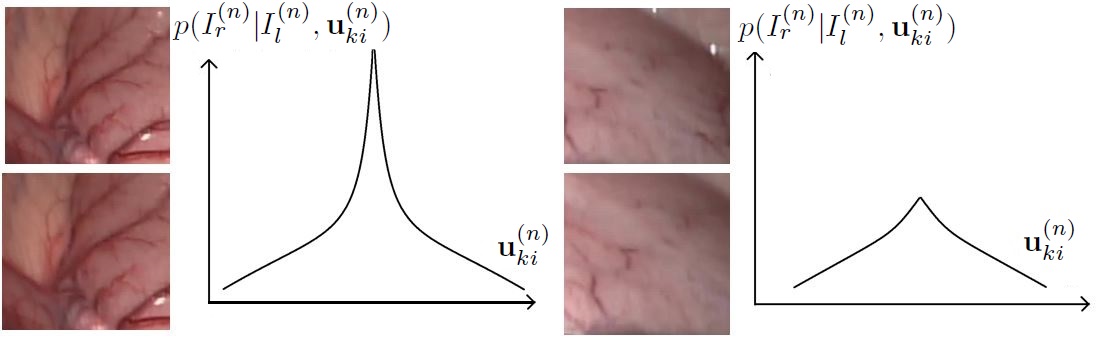}
			\end{minipage}
		}
		\caption{Presented is the patch-wise probability density function of the textureless region. It shows different probabilities (response) regarding the value of disparity on the textured and textureless tissue surface.}
		\label{fig_prob_density}
	\end{figure}
	
	\subsection{The CRFs based patch-wise posterior probability}
	\label{section_2_2}
	\sjw{The weighting strategy~\eqref{Eq_residual_fusion} in DIS~\cite{kroeger2016fast} cannot handle the fusion process well in MIS stereo images. The deterministic weights in~\eqref{Eq_residual_fusion} do not correctly depict the confidences of MIS predictions due to the textureless surface, non-Lambertian reflectance, and dark illumination.} First, the textureless surface and dark illumination in MIS lower the contrast and increase the blurriness of the obtained stereo images. As fast LK is based on the photometric consistency presumption, the parallax searching in~\eqref{Eq_fast_inverse_search} has a higher risk of falling into a local minimum. Consequently, a small residual is insufficient to measure the performance of the disparity searching. The residuals in Fig.~\ref{fig_prob_density} are both small, but the stereo matching on the textureless surface (the right one) is much less reliable. Furthermore, the photometric consistency presumption is also seriously violated on the surface affected heavily by the non-Lambertian reflectance. \sjw{Ambiguities arise when the photo-consistency assumption is violated, e.g., at intensive reflection or very dark pixels.} Even the Lambertian reflectance also pollutes the photometric consistency because the reflected intensity is related to the incident angle~\cite{shimasaki2013generating}. In the large scale environment, affine lighting formulation~\cite{engel2015large,engel2017direct} was enforced by the SLAM systems and reported to handle the photometric inconsistency well. The modeling enforces extra affine modeling of the illumination of the target image. Nevertheless, the affine modeling cannot fully tackle the complex and severe non-Lambertian reflectance in MIS. Therefore, the weights from~\eqref{Eq_residual_fusion} are misleading. This article seeks the CRFs to depict the confidence of the patch's prediction.\par
	
	CRFs is a sequential modeling technique that presents transitional probabilities between finite states based on a well-defined distribution over observations. CRFs formulates the joint probability of the states using a single exponential formulation instead of per-state models for simplification~\cite{lafferty2001conditional}. \sjw{Given state $\mathbf{s}$ and target $\mathbf{t}$, CRFs is expressed as a Boltzmann distribution~\cite{larochelle2008classification}, which is a function to measure the probability of the state as ``state's energy''.} The probability is represented in an exponential form~\cite{quattoni2004conditional} as 
	
	\begin{equation}
	\label{Eq_CRF}
	p(\mathbf{t} \mid \mathbf{s}, \mathbf{\theta})=\frac{\exp({\Psi(\mathbf{t},\mathbf{s},\mathbf{h}; \mathbf{\theta})})}{\sum_{\mathbf{t}, \mathbf{h}} {\exp(\Psi\left(\mathbf{t}, \mathbf{s}, \mathbf{h}; \mathbf{\theta} \right))}},
	\end{equation}
	
	\noindent where $\Psi(\cdot)$ is a potential function parameterized by $\mathbf{\theta}$, while $\mathbf{h}$ is a set of hidden variables that can not be directly observed~\cite{quattoni2007hidden}. Specifically, the distant observations from the target employ low potential energy in our scenario. Hence, a proper potential function follows the condition that its output is inversely relevant to the distance metric.\par
	
	\sjw{Since there is no prior knowledge of the uncertainty distribution of the left image patch and right image, it is difficult to infer the posterior probability in terms of disparity directly.} Inspired by CRFs, we choose to implicitly infer the probability with Bayesian modeling using CRFs~\cite{uzunbas2016efficient}. The posterior probability of the patch-wise disparity $\mathbf{u}_k^{(n)}$ on level $n$ is

	\begin{equation}
	\label{Eq_post_probability_0}
	\begin{aligned}
	p(\mathbf{u}_k^{(n)}|I^{(n)}_l,I^{(n)}_r)
	&\propto\frac{p(I^{(n)}_r|I^{(n)}_l,\mathbf{u}_k^{(n)})}{p(I^{(n)}_r,I^{(n)}_l,\mathbf{u}_k^{(n)})}\\
	&\propto\frac{p(I^{(n)}_r|I^{(n)}_l,\mathbf{u}_k^{(n)})}{\Sigma_{\mathbf{u}_{ki}^{(n)} \in \mathcal{P}} \ p(I^{(n)}_r|I^{(n)}_l,\mathbf{u}_{ki}^{(n)})},
	\end{aligned}
	\end{equation}
	
	\noindent where $\mathcal{P}$ is the window of all possible choice of $\mathbf{u}_{ki}^{(n)}$. In general, naturally picking a $\mathbf{u}_{ki}^{(n)}$ among $\mathcal{P}$ can be regarded as an event of equal probability. Hence, the chance of each $\mathbf{u}_{ki}^{(n)}$ to be chosen is a fixed constant. Denote $p(\mathbf{u}_k^{(n)}|I^{(n)}_l,I^{(n)}_r)$ as the probability calculated in level $n \in \mathcal{N}$ where $n$ is the pixel size on level $n$. Specifically, one pixel on level $n$ represents a $2^n \times 2^n$ pixel patch on the original image. \sjw{The computation on~\eqref{Eq_post_probability_0} is time-consuming, and we approximate it with a small window as}
	
	\begin{equation}
	\label{Eq_post_probability}
	\begin{aligned}
	&p(\mathbf{u}_k^{(n)}|I^{(n)}_l,I^{(n)}_r)\propto\frac{p(I^{(n)}_r|I^{(n)}_l,\mathbf{u}_k^{(n)})}{\mathbf{P}'+\mathbf{P}''}\propto\\
	&\frac{p(I^{(n)}_r|I^{(n)}_l,\mathbf{u}_k^{(n)})}{\mathbf{P}'}\frac{\mathbf{P}'}{\mathbf{P}'+\mathbf{P}''}\propto\frac{p(I^{(n)}_r|I^{(n)}_l,\mathbf{u}_k^{(n)})}{\mathbf{P}'}\mathrm{r}^{(n)},
	\end{aligned}
	\end{equation}
	
	\noindent where $\mathbf{P}'=\Sigma_{\mathbf{u}_{ki}^{(n)} \in \mathcal{P}'} \ p(I^{(n)}_r|I^{(n)}_l,\mathbf{u}_{ki}^{(n)})$ and $\mathbf{P}''=\Sigma_{\mathbf{u}_{ki}^{(n)} \in \mathcal{P}''} \ p(I^{(n)}_r|I^{(n)}_l,\mathbf{u}_{ki}^{(n)})$. $\mathcal{P}'$ is the small window adopted and $\mathcal{P}''$ is the rest window ($\mathcal{P}=\mathcal{P}'\cup\mathcal{P}''$).~\eqref{Eq_post_probability} simplifies $\mathcal{P}$ to a small window $\mathcal{P}'$ considering the rest candidates are numerically trivial. The compensation ratio $\mathrm{r}^{(n)} \triangleq \frac{\mathbf{P}'}{\mathbf{P}'+\mathbf{P}''}$ and is close to $1$. Since probability is assumed to follow the Gaussian distribution, the marginal patches in $\mathcal{P}''$ are small numerically. Our experiment shows that $\mathrm{r}$ ranges from $0.99-1$ and thus can be regarded as constant.\par

	Equation~\eqref{Eq_post_probability} reveals that the posterior probability of the disparity prediction can be retrieved by traversing the prior probability on all possible positions $\mathbf{u}_{ki}^{(n)}$ in the window $\mathbf{P}'$ with size $\mathrm{s}$. In addition to CRFs, we present another straightforward explanation. The direct posterior probability measurement, such as the inverse residual~\eqref{Eq_residual_fusion}, is unavoidable to suffer from the textureless surface, non-Lambertian reflectance, and dark illumination. In contrast, the posterior probability implicitly inferred from the prior probability is robust to these factors because the impact of these issues is consistent on all patches and can be compensated on the prior probabilities within the small window. Therefore, we model the prior probability $p(I^{(n)}_r|I^{(n)}_l,\mathbf{u}_{ki}^{(n)},\mathbf{x})$ based on the Boltzmann distribution, which is 
	
	
	\begin{equation}
	\label{Eq_prior_probability}
	p(I^{(n)}_r|I^{(n)}_l,\mathbf{u}_{ki}^{(n)}) = \exp\left(-\frac{\lVert I^{(n)}_l(\mathbf{u}_{ki}^{(n)})-I^{(n)}_r(\mathbf{u}_{ki}^{(n)})\lVert^2}{2{\sigma_r^{(n)}}^2\mathrm{s}^2}\right),
	\end{equation}

	\noindent where scale $\sigma_r^{(n)}$ is the hyper-parameter. Different from our preliminary work~\cite{song2021bayesian} which uses an arbitrary value, \sjw{$\sigma_r^{(n)}$ is set as the standard deviation of $\lVert I^{(n)}_l(\mathbf{u}_{k}^{(n)})-I^{(n)}_r(\mathbf{u}_{k}^{(n)})\lVert^2_\mathrm{F}$.} This dynamic parameter ensures the scale is reasonable within the window. \sjw{Results show it helps improving the accuracy and is absence of parameter tuning.} Therefore, the relative posterior probability can be obtained with~\eqref{Eq_post_probability} and~\eqref{Eq_prior_probability}.  \par

	Fig.~\ref{fig_prob_density} elaborates on the benefit of the implicit inference of the posterior probability. \sjw{For the well-textured and well-lighted image pairs}, contrary to the deterministic inverse residual weight~\eqref{Eq_residual_fusion}, the residual has strong responses (large inverse residual) at the correct parallax while the small response in another parallax. \sjw{For the textureless/dark/highlighted image pairs}, however, the response curve has smaller contrast, and the weight from~\eqref{Eq_residual_fusion} is always large in the residual. Therefore,~\eqref{Eq_residual_fusion} assigns similar high confidence to both situations.~\eqref{Eq_post_probability} solves this problem with the implicit inference and concludes that the former situation has much higher confidence than the latter. Moreover, CRFs also tests the local convergence to filter the Saddle points. The predictions identified as the Saddle points are discarded. \sjw{It should be noticed that, theoretically, this principle does not apply to small patches in highlight (a few pixels) because both large-scale and small-scale disturbances identify the small highlighted patch as well-textured. The small indistinguishable highlighted patch is mixed with texture. This issue cannot be handled unless some handcrafted threshold or prior-based method is applied. Nevertheless, we can hardly identify this issue in the experiments.}\par

	\subsection{The sGMM-based spatial Gaussian probability}
	\label{section_2_3}
	
	On top of the posterior probability quantification for the entire patch in Section~\ref{section_2_2}, \sjw{we propose a probabilistic patch kernel using spatial sGMM~\cite{reynolds2009gaussian} to estimate pixel-wise probability within the patch.} \par
	
	Theoretically, any complex distribution can be decomposed and fit with a set of simple Gaussian distributions. The target distribution can be represented as a Gaussian mixture, which is the superposition of several Gaussian kernels. \sjw{Specifically, the GMM is a probability density function in the weighted sum of Gaussian processes~\cite{reynolds2009gaussian} as} 
	
	\begin{equation}
	p(\mathbf{x} \mid \lambda)=\sum_{i} w_{i} g\left(\mathbf{x} \mid \boldsymbol{\mu}_{i}, \boldsymbol{\sigma}_{i}\right),
	\label{Eq_kernel}
	\end{equation}
	where $w_{i}$ is the weight of the kernel $g\left(\mathbf{x} \mid \boldsymbol{\mu}_{i}, \boldsymbol{\sigma}_{i}\right)$. For a balanced observation, all the $w_{i}$ can be set as the same value.
	
	\sjw{GMM consists of various Gaussian kernels with diverse parameters. A unified form of each Gaussian density kernel~\cite{rasmussen1999infinite} can be expressed as the Gaussian kernel}
	
	\begin{equation}
	g\left(\mathbf{x} \mid \boldsymbol{\mu}_{i}, \boldsymbol{\sigma}_{i}\right)=\frac{1}{\boldsymbol{\sigma_{i}}\sqrt{2 \pi}} \exp \left(-\frac{(\mathbf{x}-\boldsymbol{\mu_{i}})^{2}}{2\boldsymbol{\sigma_{i}}}\right),
	\end{equation}
	\noindent where $\boldsymbol{\mu}_{i}$ and $\boldsymbol{\sigma_{i}}$ are the mean and variance in the kernel. 
	
	\sjw{The authentic GMM is simplified as sGMM, which directly uses the prior knowledge of the parameters. \citet{jiang2018gaussian} simplified GMM by assigning $\boldsymbol{\sigma_{i}}$ with a constant and only estimated the non-exponential coefficient of Gaussian kernels. Differently, \citet{ge2015non} simplified GMM by fixing the weights and only estimate a unified $\boldsymbol{\sigma}$ to all $\boldsymbol{\sigma_{i}}$. We push them further by treating $\boldsymbol{\sigma_{i}}$ as hyperparameter while setting all weights as $1$. Thus, to avoid misunderstanding, we term it sGMM.}\par

	\sjw{In MIS, the images collected are natural and fit the presumption in sGMM.} Hence, a multivariate Gaussian distribution is adopted to measure the confidence of the pixel-wise probability using a Gaussian mask similar to~\eqref{Eq_kernel}. Considering that the confidence is measured patch-wisely, the patch's center accumulates a higher probability than the marginal pixels since the multivariate Gaussian distribution is enforced on the patch. The central pixels preserve more information. Assuming all pixels in the patch are i.i.d., we have

	\begin{equation}
	\label{Eq_spatial_prob}
	\begin{aligned}
	&p(\mathbf{u}_k^{(n)}|I^{(n)}_l,I^{(n)}_r,\mathbf{x}) \propto \\
	&p(\mathbf{u}_k^{(n)}|I^{(n)}_l,I^{(n)}_r)\sum_{\mathbf{x}_i \in \mathbf{\xi}}\exp\left(-\frac{\lVert\mathbf{x}-\mathbf{x}_i\lVert^2_\mathrm{F}}{2\sigma_s^2}\right),
	\end{aligned}
	\end{equation}

	\noindent where $\mathbf{\xi}^{(k)}(\mathbf{x})$ is the set of all pixel positions within the patch $k$ in the image coordinate. $\sigma_s$ is the 2D spatial variance of the probability in~\eqref{Eq_spatial_prob}. Note that~\eqref{Eq_spatial_prob} is independent of the patch and can therefore be pre-computed before the process. Combining~\eqref{Eq_post_probability},~\eqref{Eq_prior_probability} and~\eqref{Eq_spatial_prob}, the final pixel-wise posterior probability distribution on level $n$ can be represented as 
	
	\begin{equation}
	\label{Eq_final_prob}
	\begin{aligned}
	&p(\mathbf{u}_k^{(n)}|I^{(n)}_l,I^{(n)}_r,\mathbf{x}) \propto 
	\sum_{\mathbf{x}_i \in \mathbf{\xi}}\exp\left(-\frac{\lVert\mathbf{x}-\mathbf{x}_i\lVert^2_\mathrm{F}}{2\sigma_s^2}\right) \\
	&\frac{\exp\left(-\frac{\lVert I^{(n)}_l(\mathbf{u}_{k}^{(n)})-I^{(n)}_r(\mathbf{u}_{k}^{(n)})\lVert^2_\mathrm{F}}{2{\sigma_r^{(n)}}^2\mathrm{s}^2}\right)}
	{\sum_{\mathbf{u}_{ki}^{(n)} \in \mathcal{P}}\exp\left(-\frac{\lVert I^{(n)}_l(\mathbf{u}_{ki}^{(n)})-I^{(n)}_r(\mathbf{u}_{ki}^{(n)})\lVert^2_\mathrm{F}}{2{\sigma_r^{(n)}}^2\mathrm{s}^2}\right)}.
	\end{aligned}
	\end{equation}  
	
	To avoid misunderstanding, we should emphasize that~\eqref{Eq_prior_probability} and~\eqref{Eq_spatial_prob} are not the cost functions but probability functions for each patch or pixel. The probability quantifies the weight for the fusion process. Our probabilistic Bayesian method does not require heavy computational cost in the optimization.
	
	\subsection{Probability propagation from coarse-to-fine}
	\label{section_2_4}
	In addition to the Bayesian probability in~\eqref{Eq_post_probability} at one level, \sjw{the measured probabilities on the coarser levels should also be propagated to this level since the probability should describe multi-scale observations.} The probability on the finest level should count the probability of all scales. However, as the estimated disparity from the coarser level is only used as initialization for the finer level inverse searching,  the exact form of probability propagation cannot be accurately obtained. We approximate it by treating the disparity estimation on each level as independent processes with different weights. This is reasonable because the inverse searching on each level is bounded, and the coarsest prediction does not deviate much from the final prediction. Therefore, all scales contribute to the searching of the finest disparity unevenly.\par

	We approximate it with an empirical formulation. Define $\Gamma(p(\mathbf{u}_k^{(n)}|I^{(n)}_l,I^{(n)}_r))$ as the probability contributed from level $n$. The probability which considers coarser levels can be obtained as 
	
	\begin{equation}
	\label{Eq_bayesian_coarselevel}
	\begin{aligned}
	p(\mathbf{u}_k^{(n)}|I^{(n)}_l,I^{(n)}_r,\mathbf{x}) \propto \sum_{n\in\mathcal{N}}\Gamma(p(\mathbf{u}_k^{(n)}|I^{(n)}_l,I^{(n)}_r,\mathbf{x})), 
	\end{aligned}
	\end{equation}  
	
	\noindent where $\mathcal{N}$ is the set of all levels. In this paper, $\Gamma(p(\mathbf{u}_k^{(n)}|I^{(n)}_l,I^{(n)}_r))$ is approximated as the following because the finer scale is $2$ times larger in one dimension, which is 
	
	\begin{equation}
	\begin{aligned}
	\Gamma(p(\mathbf{u}_k^{(n)}|I^{(n)}_l,I^{(n)}_r,\mathbf{x})) = \frac{2^n}{\sum_{n\in\mathcal{N}}2^n}p(\mathbf{u}_k^{(n)}|I^{(n)}_l,I^{(n)}_r,\mathbf{x}). 
	\end{aligned}
	\end{equation} 
	It is worth noting that any successful observation at an arbitrary level can lead to an effective observation from global perspective. Hence, the formulated probability is in the form of superposition to represent such logical ``or'' relationship.\par
	
	The pixel-wise weighting in~\eqref{Eq_residual_fusion} is substituted with \par
	
	\begin{equation}
	\label{Eq_residual_fusion_BDIS}
	\hat{\mathbf{u}}_\mathbf{x}^{(n)} = \sum_{k \in \Omega} \Gamma(p(\mathbf{u}_k^{(n)}|I^{(n)}_l,I^{(n)}_r,\mathbf{x}))  \mathbf{u}_k^{(n)}.
	\end{equation}
	
	Based on the fused disparity in the finest scale $\hat{\mathbf{u}}_\mathbf{x}^{(f)}$($(f)$ is the finest scale), the depth can be obtained following the routine stereo vision process, which is\par

	\begin{equation}
	\label{Eq_depth_fusion}
	\mathrm{d}_\mathbf{x} = \frac{\mathrm{f}\mathrm{b}}{\hat{\mathbf{u}}_\mathbf{x}^{(f)}},
	\end{equation}
	\noindent where $\mathrm{f}$ and $\mathrm{b}$ are the focal length and baseline.  \par

	\subsection{PDF estimation of the depth}
	\label{section_var}
	
	This section provides an optional inaccurate depth variance estimation method in addition to the workflow shown in Fig. \ref{fig_framework}. Although~\eqref{Eq_post_probability} obtains the patch's posterior probability $\mathbf{u}_k$, the parameter of the PDF, specifically the variance, is unknown. With the Gaussian noise assumption, confidence quantification plays an important role in 3D geometry and sensor fusion. \sjw{Assuming the errors are predominated by Gaussian distribution}, we push~\eqref{Eq_post_probability} toward the estimation of the PDF. PDF estimation technique~\cite{PRML} is employed and modified. Denote the parameter set as $\theta_k = \{\mathbf{u}_k,\sigma_k\}$, where $\sigma_k$ is the standard deviation of the patch $k$. Following the authentic definition of the MAP algorithm, $\theta_k$ can be obtained as
	
	\begin{equation}
	\label{Eq_MAP_authentic}
	\begin{aligned}
	\theta_{k} &=\arg \max _{\theta_k} p(\theta_k \mid \mathbf{P}')=\arg \max _{\theta_k} \frac{p(\mathbf{P}' \mid \theta_k) p(\theta_k)}{p(\mathbf{P}')} \\
	&=\arg \max _{\theta_k} p(\mathbf{P}' \mid \theta_k) p(\theta_k)\\
	&=\arg \max _{\theta_k} \sum_{i\in \mathbf{P}'}\ln p\left(\mathbf{u}_{ki}^{(n)} \mid \theta_k\right)+\ln p(\theta_k),
	\end{aligned}
	\end{equation}
	
	\noindent where $p(\theta_k)$ is the prior distribution of the parameter $\theta_k$. $p\left(\mathbf{u}_{ki}^{(f)} \mid \theta_k\right)$ is equivalent to the distribution of $p(I_r^{(f)}|I^{(f)}_l,\mathbf{u}_{ki}^{(f)})$ on condition of the PDF's parameter $\theta_k$. The MAP formulation~\eqref{Eq_MAP_authentic} can be applied in BDIS after three modifications. First, the prior distribution of $\mathbf{u}_k$ is fixed at the fast LK's estimation $\mathbf{u}_k^{(f)}$. Second, the prior knowledge of the distribution of $\sigma_k$ is unknown and we do not enforce additional prior knowledge term. Lastly,~\eqref{Eq_post_probability} indicates an approximate ratio $\mathrm{r}^{(f)}$, an extra state $\mathrm{c}_k$ helps compensating the impact of inaccuracy resulted from the small processing window $\mathbf{P}'$. Applying the three modifications in MAP process, we have 
	
	\begin{equation}
	\label{Eq_MAP_1}
	\begin{aligned}
	&\min_{\sigma_k,\mathrm{c}_k}  \sum_{i\in \mathbf{P}'}\frac{\mathrm{c}_k}{\sigma_k \sqrt{2 \pi}} \exp{\left(-\frac{(\mathbf{u}_{ki}^{(f)}-\mathbf{u}_k^{(f)})^{2}}{2 \sigma_k^{2}}\right)}\\
	&-p(I^{(f)}_r|I^{(f)}_l,\mathbf{u}^{(f)}_{ki}).
	\end{aligned}
	\end{equation}
	
	To enable efficient optimization of~\eqref{Eq_MAP_1}, natural logarithm function $\ln(\cdot)$ is applied.
	
	\begin{equation}
	\label{Eq_MAP_2}
	\begin{aligned}
	&\min_{\sigma_k,\mathrm{c}_k} \sum_{i\in \mathbf{P}'} \ln(\mathrm{c}_k)-\frac{1}{2}\ln\left(\sigma_k^2 2\pi\right)-\frac{(\mathbf{u}_{ki}^{(f)}-\mathbf{u}_k^{(f)})^{2}}{2\sigma_k^2}\\
	&-\ln(p(I_r|I_l,\mathbf{u}_{ki}^{(f)})).
	\end{aligned}
	\end{equation}

	Gauss-Newton (GN) algorithm is adopted to solve~\eqref{Eq_MAP_2}. \sjw{The linearized system is solved with Cholesky decomposition for fast solving.} The initial input of $\sigma_k$ and $\mathrm{c}_k$ are set as $\sqrt{0.1}$ and $1$. To reveal the reliability of GN in solving~\eqref{Eq_MAP_2}, massive amount of Monte-Carlo experiments were conducted. \sjw{The experiment indicates~\eqref{Eq_MAP_2} has a local minimum if $\sigma_k$ is set much larger than the reference.} Thus, the initial $\sigma_k$ is set as $\sqrt{0.1}$, which is enough because the variance in practice should be much larger than $\sqrt{0.1}$. Meanwhile, the convergence of~\eqref{Eq_MAP_2} with GN is very fast. Even though 5 iterations are enough in our experiments, we fix the optimization iteration to 10. The extremely small size of the state vector and Jacobian matrix makes the computation small.\par
	
	Additionally,~\eqref{Eq_MAP_2} needs to be handled with two exceptions. First, the residual of~\eqref{Eq_MAP_2} is too large. The threshold for residual in~\eqref{Eq_MAP_2} is set as $0.1$. Second, the unconvergence in the minimization process. For example, all probabilities are almost equal, or the probability at $\mathbf{u}_k^{(f)}$ is not the largest. Both situations hint the inconsistency of the Gaussian distribution assumption. Therefore, we compromise this paradox by manually setting the detected patch with a large standard deviation which alleviates the issue. Finally, it is worth clarifying that $\sigma_x$ is assigned as the patch searching threshold in the fast LK process. This is reasonable because the patch's probability (and weight) is small in both situations. \sjw{Besides, it should be emphasized that although some PDFs are inconsistent with the Gaussian distribution presumption, the proportional probability~\eqref{Eq_final_prob} is still valid.} \par
	
	The patch-wise variance $\sigma_k^2$ is propagated to pixel-wise depth $\sigma_x^2$ (at position $\mathbf{x}$) following weights defined in~\eqref{Eq_final_prob} and the disparity to depth process defined in~\eqref{Eq_residual_fusion_BDIS}. The propagation function is 
	
	\begin{equation}
	\label{Eq_var_depth}
	\sigma_{x}^2 = \frac{(\mathrm{f}\mathrm{b})^2}{{\hat{\mathbf{u}}^{(f)}_\mathbf{x}}^4}\left(\sum_{k \in \Omega}\Gamma(p(\mathbf{u}_k^{(f)}|I^{(f)}_l,I^{(f)}_r,\mathbf{x}))^2 \sigma_k^2\right).
	\end{equation}
	
	It should be emphasized that the PDF estimation is partial and inaccurate for two reasons. The proportional probability~\eqref{Eq_post_probability} strictly follows the Bayesian theory and is independent of the type of distribution. The PDF estimation, however, requires the consistency of the Gaussian distribution presumption of the error. There is no guarantee for consistency. In many data sets, we do notice a large amount of probability~\eqref{Eq_prior_probability} within the window perfectly follows the Gaussian curve. Furthermore, multi-scale fast LK suffers from the local minima or bad convergence, which can be categorized as epistemic uncertainty. Thus, our uncertainty estimation method is optional and for reference only.\par
	
	\begin{figure*}[htpb]    
		\centering
		\subfloat{    
			\begin{minipage}[htpb]{0.25\textwidth}    
				\centering
				\includegraphics[width=1\linewidth]{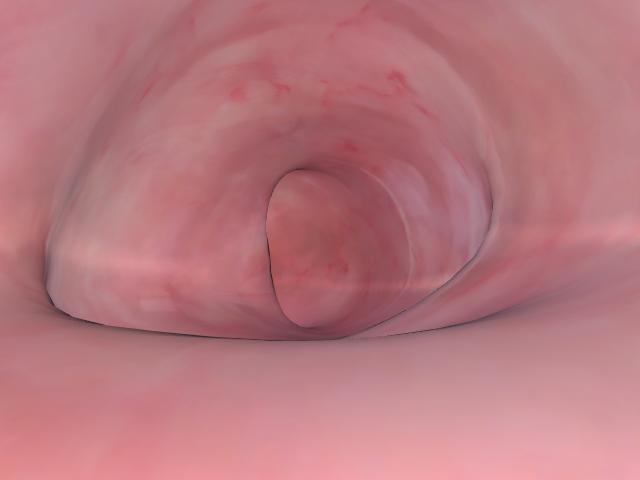}            
			\end{minipage}                
		}
		\subfloat{    
			\begin{minipage}[htpb]{0.25\textwidth}    
				\centering
				\includegraphics[width=1\linewidth]{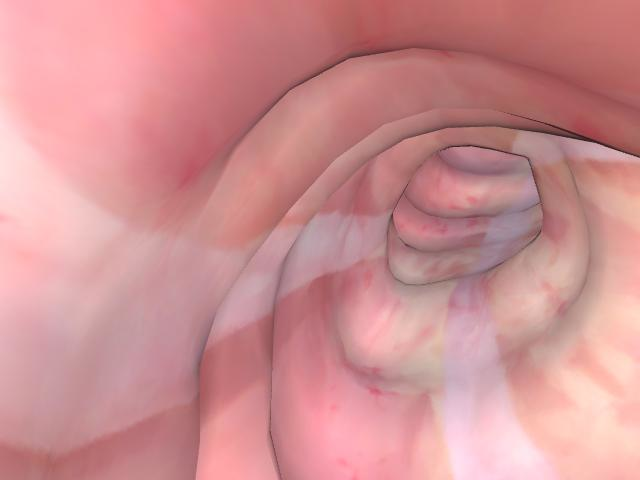}            
			\end{minipage}                
		}
		\subfloat{    
			\begin{minipage}[htpb]{0.25\textwidth}    
				\centering
				\includegraphics[width=1\linewidth]{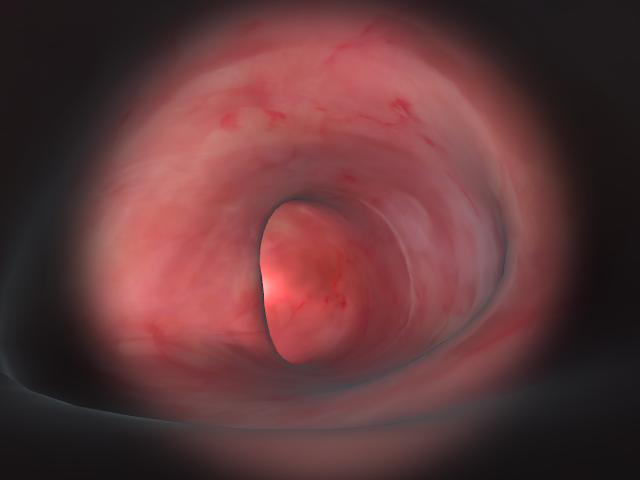}            
			\end{minipage}                
		}
		\subfloat{    
			\begin{minipage}[htpb]{0.25\textwidth}    
				\centering
				\includegraphics[width=1\linewidth]{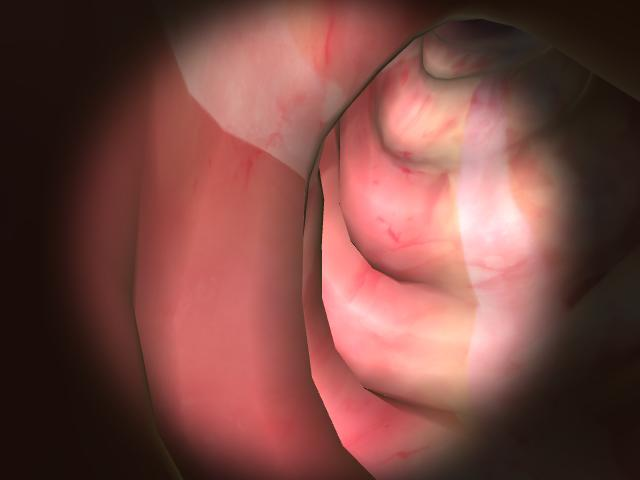}            
			\end{minipage}                
		}
		\caption{\sjw{The figure shows sample synthetic left images of both diffuse lighting and non-Lambertian reflectance.}}
		\label{fig_exvivo_smamples}
	\end{figure*}

	\begin{figure*}[htpb]    
		\centering
		\subfloat{    
			\begin{minipage}[htpb]{0.25\textwidth}    
				\centering
				\includegraphics[width=1\linewidth]{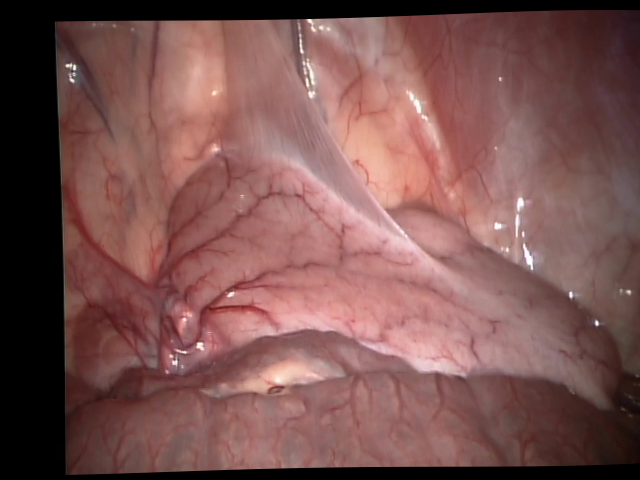}            
			\end{minipage}                
		}
		\subfloat{    
			\begin{minipage}[htpb]{0.25\textwidth}    
				\centering
				\includegraphics[width=1\linewidth]{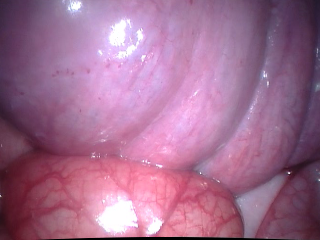}            
			\end{minipage}                
		}
		\subfloat{    
			\begin{minipage}[htpb]{0.25\textwidth}    
				\centering
				\includegraphics[width=1\linewidth]{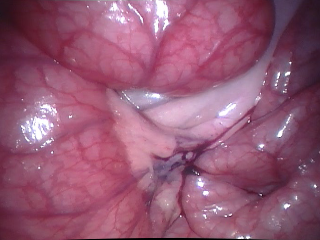}            
			\end{minipage}                
		}
		\subfloat{    
			\begin{minipage}[htpb]{0.25\textwidth}    
				\centering
				\includegraphics[width=1\linewidth]{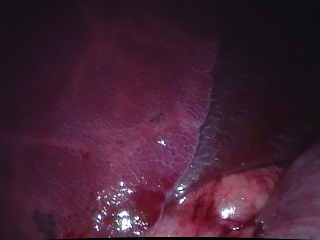}            
			\end{minipage}                
		}
		\caption{Sample left images from the Hamlyn stereo data set. }
		\label{fig_Hamlyn_samples}
	\end{figure*}
	
	\begin{figure*}[htpb]    
		\centering
		\subfloat{    
			\begin{minipage}[htpb]{0.25\textwidth}    
				\centering
				\includegraphics[width=1\linewidth]{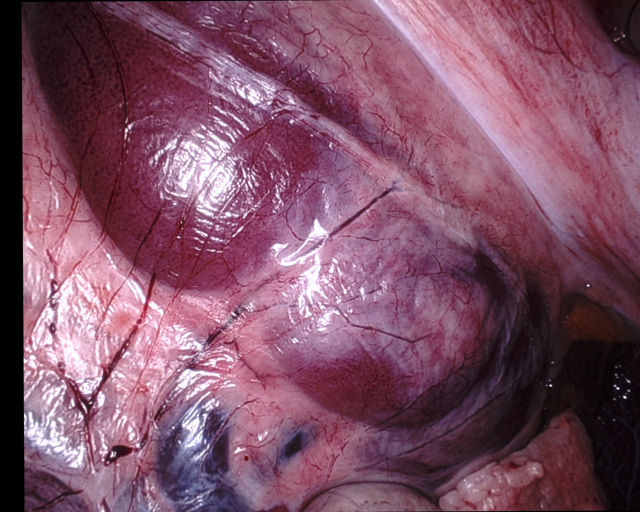}            
			\end{minipage}                
		}
		\subfloat{    
			\begin{minipage}[htpb]{0.25\textwidth}    
				\centering
				\includegraphics[width=1\linewidth]{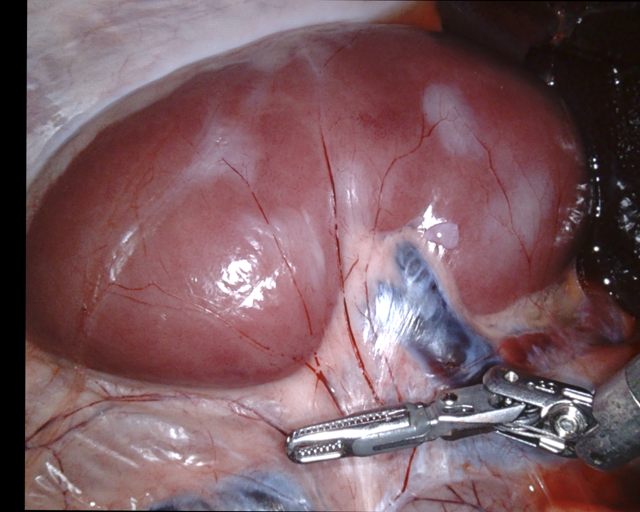}            
			\end{minipage}                
		}
		\subfloat{    
			\begin{minipage}[htpb]{0.25\textwidth}    
				\centering
				\includegraphics[width=1\linewidth]{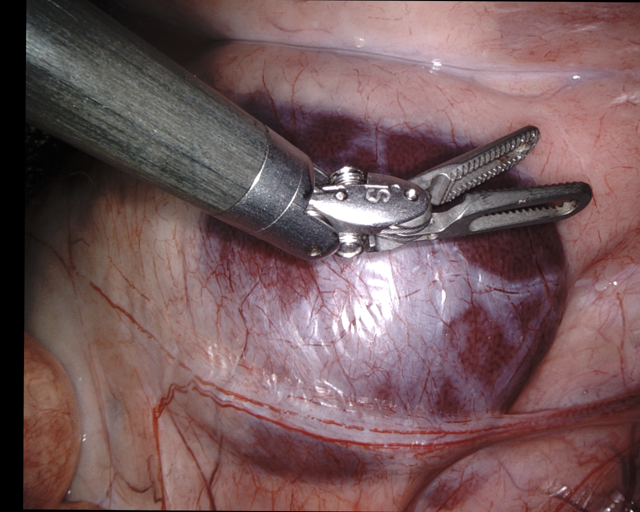}            
			\end{minipage}                
		}
		\subfloat{    
			\begin{minipage}[htpb]{0.25\textwidth}    
				\centering
				\includegraphics[width=1\linewidth]{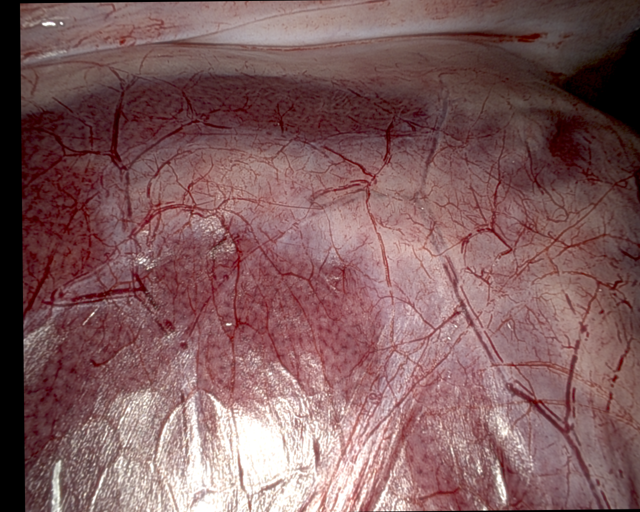}            
			\end{minipage}                
		}
		\caption{The figure shows sample calibrated left images provided by the SCARED data set. Surgical instruments are presented in two of the images.}
		\label{fig_MICCAI_samples}
	\end{figure*}

	\begin{figure*}[htpb]    
		\centering
		\subfloat{    
			\begin{minipage}[htpb]{0.25\textwidth}    
				\centering
				\includegraphics[width=1\linewidth]{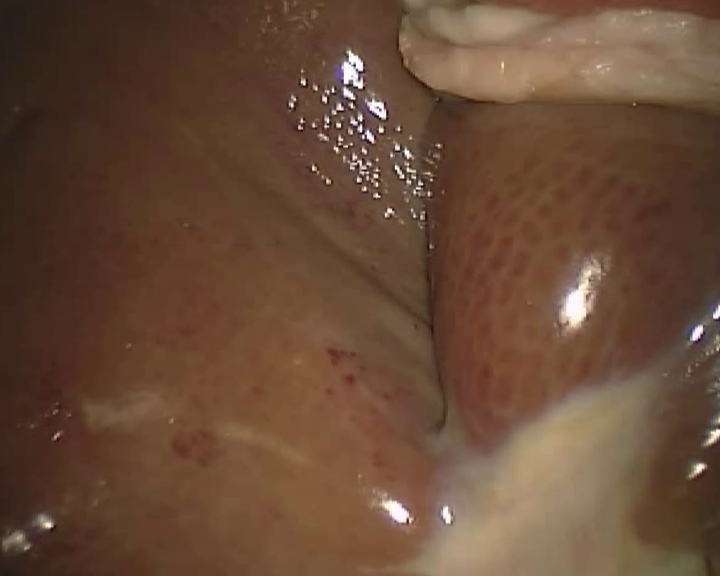}            
			\end{minipage}                
		}
		\subfloat{    
			\begin{minipage}[htpb]{0.25\textwidth}    
				\centering
				\includegraphics[width=1\linewidth]{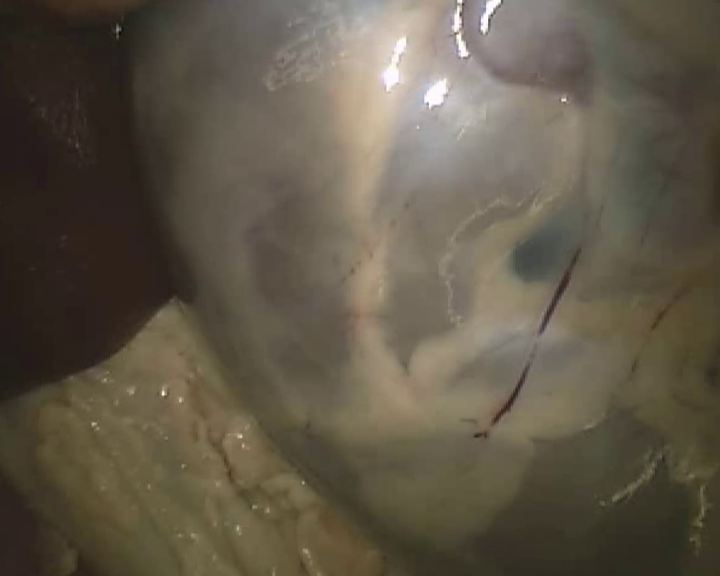}            
			\end{minipage}                
		}
		\subfloat{    
			\begin{minipage}[htpb]{0.25\textwidth}    
				\centering
				\includegraphics[width=1\linewidth]{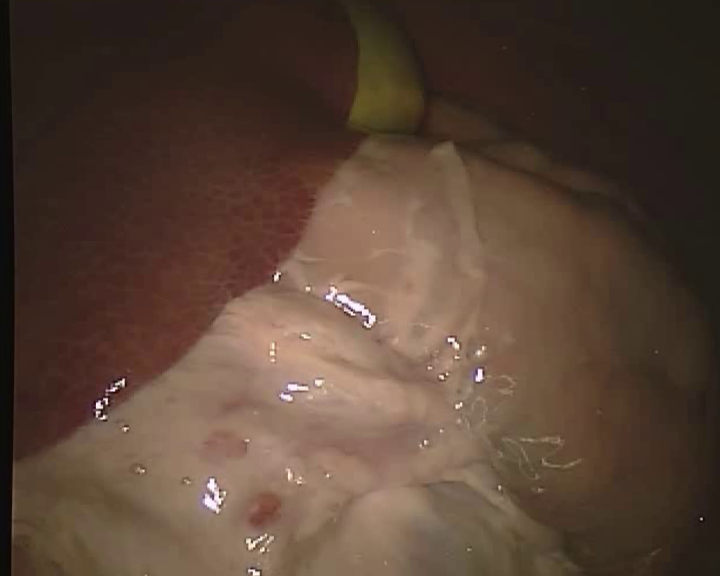}            
			\end{minipage}                
		}
		\subfloat{    
			\begin{minipage}[htpb]{0.25\textwidth}    
				\centering
				\includegraphics[width=1\linewidth]{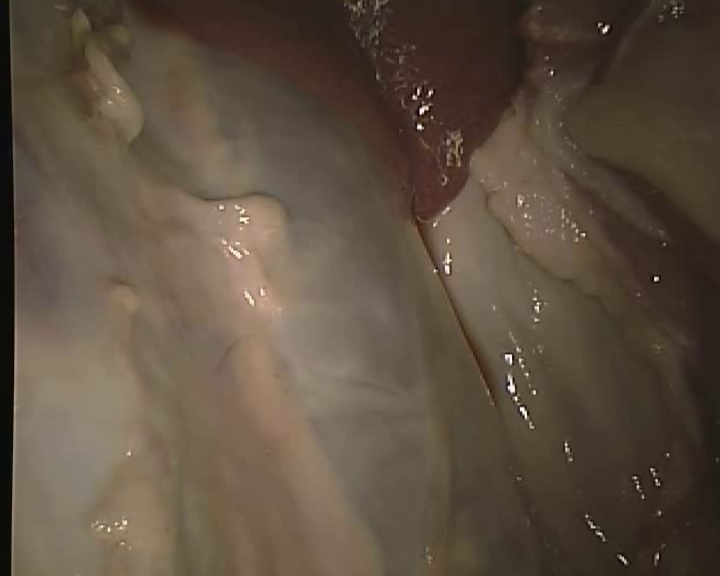}            
			\end{minipage}                
		}
		\caption{The figure shows sample calibrated left images provided by the SERV-CT stereo data set. From left to right are data set 1, 5, 9 and 15.}
		\label{fig_SERV_CT_samples}
	\end{figure*}
	
	\subsection{Technical details}
	
	This section covers important technical details in the implementation. First, in the fast LK~\eqref{Eq_fast_inverse_search} process of some patches, the algorithm stops before the convergence due to reaching the maximum iteration limit. Consequently, the unconverged parallax prediction is not the minimum in the CRFs process~\eqref{Eq_post_probability}. Once detected, the unconverged predictions are discarded. Second, some image patch has invalid pixels caused by image padding, undistortion, or rectification. We follow ELAS~\cite{geiger2010efficient} and use threshold $\gamma$ to filter small patches without enough valid predicted pixels in the final step. Third, the exact exponential computation consumes massive computational resources. \sjw{\citet{schraudolph1999fast} adopted the practical approximation method, which takes advantage of the manipulation of a standard floating-point representation. Our experiment shows the approximated exponential computation only takes less than $1/5$ computational time of the original exponential function in the standard C++ library.}~\citet{schraudolph1999fast} reported that the error is bounded for less than $4\%$ which is acceptable in our case.\par
	
	Finally, we would like to clarify that the Bayesian probability quantification method in Section~\ref{section_2_2},~\ref{section_2_3},~\ref{section_2_4} is independent of the choice of the base method or even DNN. \sjw{It is coupled with fastLK because we aim at CPU-level real-time stereo matching algorithm while preserving high accuracy.} It has the potential to be applied to other stereo matching algorithms, including the DNN-based methods. \par

	\section{Results and discussion}
	\label{section_result}
	The efficiency of BDIS was validated from the perspectives of accuracy and time consumption. This section firstly introduces the data sets. Next, the accuracy comparisons on both the synthetic data set and in-vivo data set are presented. The following content summarizes the time consumption of all the baseline methods. Lastly, an ablation study was conducted to discuss the contributions of different strategies and modules.\par
	
	\subsection{\sjw{Baseline algorithms, data sets, and metrics}}
	
	Prior-free methods DIS~\cite{kroeger2016fast}, SGBM~\cite{hirschmuller2005accurate} and ELAS~\cite{geiger2010efficient} were adopted as the baseline methods to be compared with BDIS in the in-vivo and synthetic data sets experiments \footnote{Readers are encouraged to watch the attached video. The code also provides the visualization.}. Following the recommendations from~\cite{allan2021stereo}, the DNN-based methods PSMNet~\cite{chang2018pyramid} and GwcNet~\cite{guo2019group} were also employed for comparison. The computation platform was a commercial desktop with CPU i5-9400 and GPU GTX 1080ti. DIS and SGBM are open-sourced and provided from OpenCV (C++ version). The code of ELAS is also publicly available. We implemented BDIS in C++, which was based on the code from~\cite{kroeger2016fast}. All the prior-free methods were compiled and run on the CPU end (single). The two open-sourced DNN-based methods were implemented on PyTorch~\cite{paszke2019pytorch} and run on the GPU end.\par

	\sjw{A virtual surgical data set was synthesized to control factors, including the intrinsic parameters, reference depth, and calibration. The synthesized data set ensures absolute accuracy in camera parameter calibration, reference depth retrieval, and image rectification and undistortion.} Our synthetic data set was generated from an off-the-shelf virtual phantom of a male's digestive system. The stereoscope was placed inside the colon, and we implemented a module to enable it moving automatically in the colon. The 3D game engine Unity3D\footnote{\url{https://unity.com/}} was adopted to generate the sequential RGB stereo and depth images, which strictly follows the pin-hole camera projection model. The size of all images is $640 \times 480$. The number of stereo image pairs collected from the colon is $310$ ($300$ was collected for training DNN-based methods only). For better comparison, both diffuse lighting and non-Lambertian reflectance were simulated. \sjw{It should be noticed that the dark region near the edges is in low illumination, not pure black.} Sample images are presented in Fig.~\ref{fig_exvivo_smamples}. Details of the training data (DNN-based methods only) set will be described in each experiment section.\par

	\sjw{Moreover, two in-vivo and one ex-vivo data sets were adopted for comparison. The first in-vivo comes from the public Hamlyn in-vivo stereo videos~\cite{giannarou2013probabilistic}.} We adopted 200 image pairs with size $640\times480$ and 200 image pairs with size $288\times 360$. \sjw{The Hamlyn data set does not contain the reference depth.} Sample images can be found in Fig.~\ref{fig_Hamlyn_samples}. \sjw{The second in-vivo data set (we name it SCARED) is from stereo correspondence and reconstruction of endoscopic data sub-challenge, which was organized during the Endovis challenge at the conference MICCAI 2019~\cite{allan2021stereo} (samples images are shown in Fig.~\ref{fig_MICCAI_samples}).} This data set consists of the stereo images with the annotated depth captured using a da Vinci Xi surgical robot. The size of the images is $1280 \times 720$. It has 7 training data sets and 2 testing data sets (5 keyframes for each data set). Only 5 training data sets and 2 testing data sets were used because the rest 2 are associated with incorrect calibration parameters. Keyframe 4 in data set 1 is also not correct. All the in-vivo stereo images (34 overall) were rectified, undistorted, calibrated, and vertically aligned with the provided intrinsic and extrinsic parameters. \sjw{The ex-vivo data set SERV-CT \cite{EDWARDS2022102302} was also obtained from the classic da Vinci Surgical System. SERV-CT contains 16 stereo endoscopic image pairs with reference anatomical segmentation derived from CT. Two different ex-vivo porcine samples were imaged using the straight and $30^\circ$ endoscopes. A CT scan provided the reference from the O-arm system. The scan contains both the anatomy and the endoscope, facilitating constrained manual alignment to provide the pose of the viewed anatomical surface relative to the endoscope. As sample calibrated left images in Fig. \ref{fig_SERV_CT_samples} indicate, the retrieved images in SERV-CT are less textured than SCARED data set.}\par

	We used the following parameter setting: Coarsest scale was $2^5$, and finest scale was $2^1$. min/max iterations 12, early stopping parameters 0.05 0.95 0.10, patch size 10, patch overlap 0.55, and no left to right consistency check was enforced; Patch mean normalization was enforced to reduce the impact of illumination; No M-estimator was used. $\gamma$ was set to $0.75$ for $640\times480$ and $0.25$ for $288\times 360$ data to discard the patch without enough valid pixels. The sampling within one Bayesian window was 5; the disturbance from the convergence was $0.5$ and $1$ pixel; The minimal ratio of the valid patch was set to $0.75$;\sjw{$\sigma_s$ was set to $4$. Pixel-wise threshold $0.15$ is enforced to filter unreliable predictions}. Regarding the preliminary work~\cite{song2021bayesian}, most parameters were the same except patch overlap set 0.55 instead of 0.6, allowing almost 2 times faster speed. \par
	
	\sjw{For each pixel $\mathbf{x}$ with estimated depth $\mathrm{d}_\mathbf{x}$, define the absolute depth error as} 
	
	\begin{align*}
	\label{Eq_abs_error}
	\mathrm{e}_{\mathbf{x}}=\lVert \mathrm{d}_\mathbf{x} - \overline{\mathrm{d}_\mathbf{x}} \lVert,
	\end{align*} 
	
	\noindent \sjw{where $\overline{\mathrm{d}_\mathbf{x}}$ is the reference depth at pixel $\mathbf{x}$.} For each image, we count the mean and median errors in the entire image. For the data set with multiple images, the metrics ``average mean error'' and ``average median error'' refer to the average of mean and median errors in the data set.\par
	
    \sjw{The numbers of predicted pixels were counted as ``validity'' or ``valid pixels'' in the experiments. For robustness, some predicted pixels' disparity are ignored by ELAS, SGBM, and BDIS. It is reasonable to discard  relatively small regions that is believed to be not confident.}\par

	\begin{table*}[]
		\centering
		\setlength\tabcolsep{2pt} 
		\caption{The table is the average mean and median absolute depth error comparisons on the synthetic colon data with diffused light and non-Lambertian reflectance shown in Fig.~\ref{fig_exvivo_smamples}. Results of the prior-free methods ELAS, SGBM, DIS, and the proposed BDIS are presented. Prior-based methods GwcNet and PSMNet are also shown. ``Median'' means the median error. ``Mean'' refers to mean error. ``Validity'' is the number of valid predicted pixels. The error is presented in mm. The number of valid pixels is $1000$. BDIS$^\triangle$ refers to our preliminary work~\cite{song2021bayesian}. \sjw{``Time'' refers to time consumption measured in second.}}
		\sjw{\begin{tabular}{p{1.15cm}<{\raggedright}|p{0.91cm}<{\centering}|p{0.98cm}<{\centering}|p{0.93cm}<{\centering}|p{1.21cm}<{\centering}|p{1.21cm}<{\centering}|p{1.11cm}<{\centering}|p{0.93cm}<{\centering}|p{0.91cm}<{\centering}|p{0.93cm}<{\centering}|p{0.98cm}<{\centering}|p{1.21cm}<{\centering}|p{1.21cm}<{\centering}|p{1.11cm}<{\centering}|p{0.93cm}<{\centering}}
			\toprule 
			& \multicolumn{7}{c|}{Diffuse light}                                                                                     & \multicolumn{7}{c}{Non-Lambertian reflectance}                                             \\ \midrule \midrule
			& ELAS   & SGBM   & DIS    & GwcNet & PSMNet & BDIS$^\triangle$                     & BDIS                            & ELAS  & SGBM  & DIS    & GwcNet & PSMNet & BDIS$^\triangle$   & BDIS                            \\ \midrule \midrule
			Median         & 0.178  & 0.512  & 0.251  & 0.542      & 0.417      & \textbf{0.161} & \textbf{0.161} & 0.157 & 1.211 & 0.338  & 0.248      & 0.528      & 0.142  & \textbf{0.126} \\
			Mean        & 0.220  & 1.113  & 0.753  & 0.809      & 0.641      & 0.320                           & \textbf{0.202} & 0.295 & 1.660 & 0.702  & 0.475      & 0.850      & 0.316  & \textbf{0.221} \\
			Validity & 166.77 & 103.92 & 288.41 & 100.00     & 301.42     & 208.44                          & 167.07                        & 153.00 & 115.19 & 274.91 & 143.20     & 303.28    & 226.03 & 142.05                          \\ 
			Time & 0.256 & 0.113 & 0.036 & 0.286     & 0.375     & 0.086 & 0.066     
			& 0.221 & 0.094 & 0.032 & 0.281     & 0.366    & 0.082 & 0.061                            \\ \bottomrule
		\end{tabular}
		}
		\label{Table_exvivo_dataset_colon}
	\end{table*}

	\begin{figure*}[htpb]    
		\centering
		\subfloat{    
			\begin{minipage}[htpb]{1\textwidth}    
				\centering
				\includegraphics[width=1\linewidth]{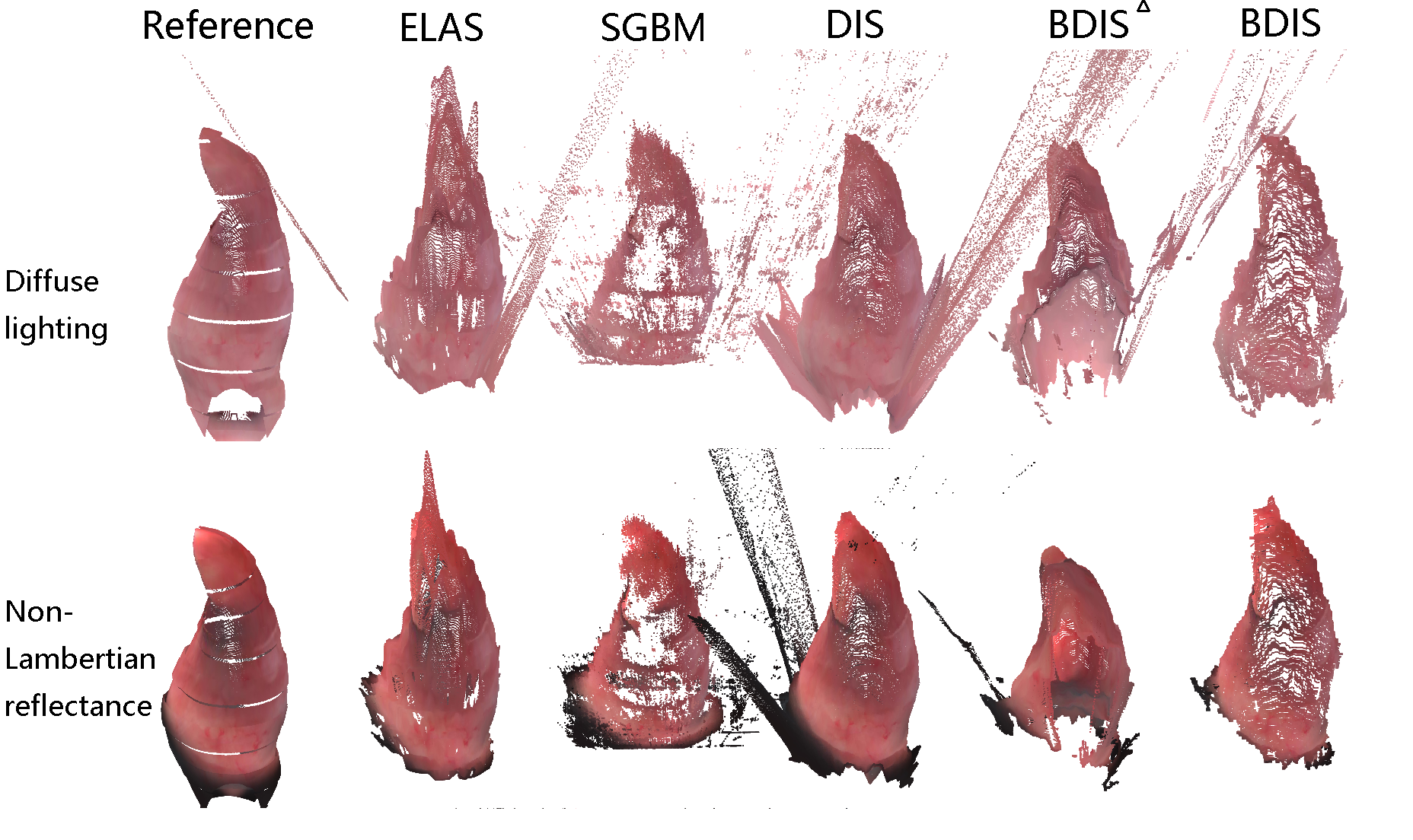}            
			\end{minipage}                
		}\\
		\caption{\sjw{The figure shows the sample reconstructions in diffuse lighting and non-Lambertian reflectance scenarios. BDIS$^\triangle$ refers to our preliminary work~\cite{song2021bayesian}. These correspond to the diffused lighting images shown in Fig.~\ref{fig_exvivo_smamples}. Readers are encouraged to watch the attached video for more restuls.}
		\label{fig_exvivo_comparison_1}}
	\end{figure*}

	\subsection{Comparisons on the synthetic data set}
	\label{section_synthetic}
	
	The proposed BDIS was first compared with the prior-free methods ELAS, SGBM, DIS, and our preliminary work~\cite{song2021bayesian} on the synthesized data set. \sjw{Moreover, DNN-based methods PSMNet~\cite{chang2018pyramid} and GwcNet~\cite{guo2019group} were also tested following the recommendation of \citet{allan2021stereo}. They were first trained on the $300$ additionally synthesized training image pairs in the virtual colon. The $300$  labeled pairs were employed to fine-tune the pre-trained network, which was obtained by training on the city-scape Kitti2015 data sets~\cite{geiger2013vision,Menze2018JPRS}}. In the fine-tuning process, $260$ and $40$ were used for training and validating, respectively, in the colon data set. Both approaches were trained with the Adaptive movement estimation (Adam) optimizer~\cite{kingma2014adam}. The fine-tuned training was conducted with 300 epochs of training. The rest parameter settings of the DNN-based methods strictly followed the settings in their original works~\cite{chang2018pyramid,guo2019group}. The quantitative accuracy comparisons of the colon data set are presented in Table~\ref{Table_exvivo_dataset_colon} in the average mean and median absolute depth error. \sjw{The synthetic data set is free of inaccurate stereo image calibration and reference depth.} Fig.~\ref{fig_exvivo_comparison_1} presents the sample reconstructions of the baseline approaches. Table~\ref{Table_exvivo_dataset_colon} suggests DNN-based methods do not perform well on the synthetic data sets. \par

	\begin{figure*}[htpb]    
		\centering
		\subfloat{    
			\begin{minipage}[htpb]{1\textwidth}    
				\centering
				\includegraphics[width=1\linewidth]{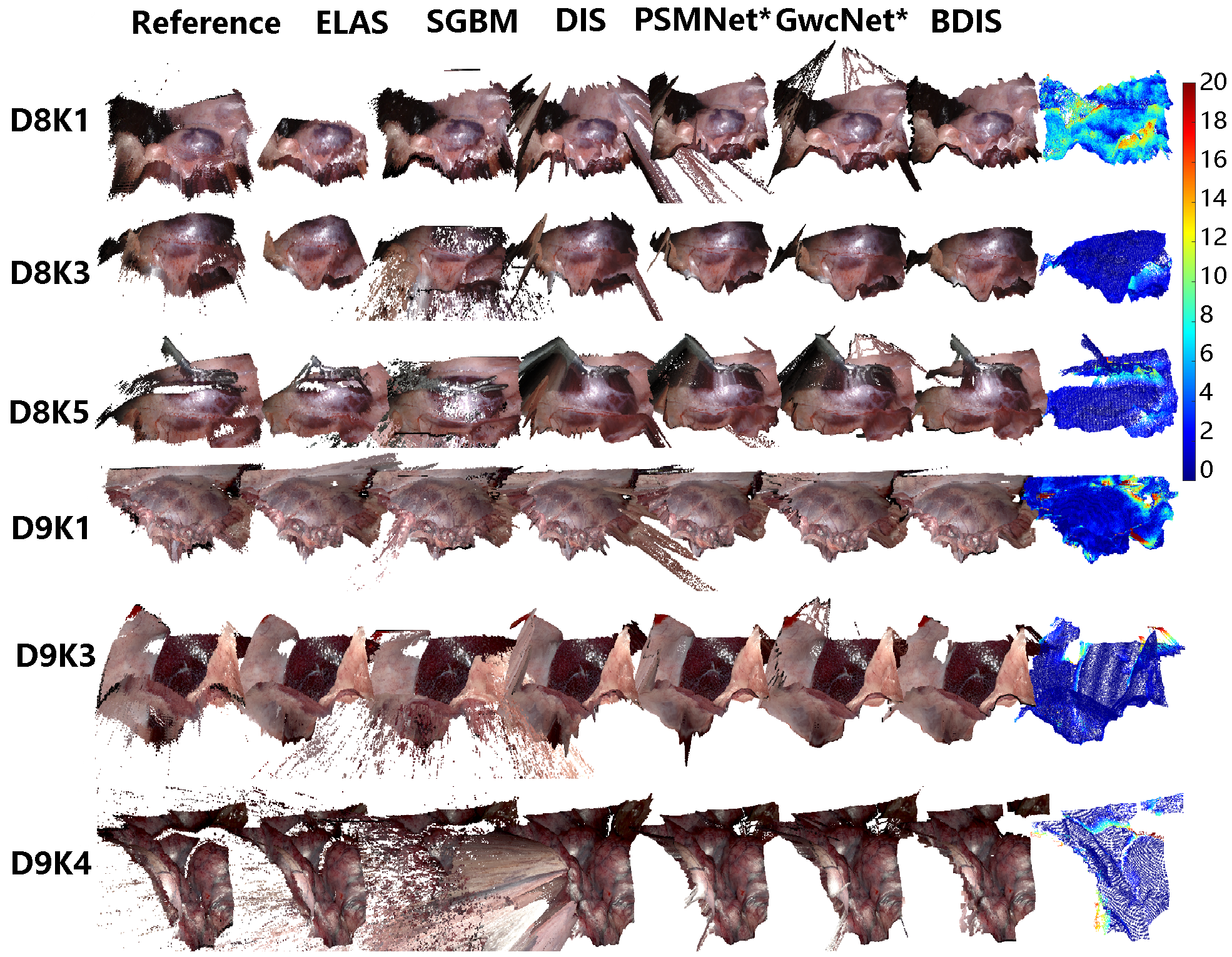}            
			\end{minipage}                
		}
		\caption{\sjw{The figure shows sample reconstructions of the SCARED stereo data set. D\textit{X}K\textit{Y} means the keyframe \textit{Y} in data set \textit{X}. The error map (in mm) of the last column is the error of BDIS. Readers are encouraged to refer to the attached video to appreciate the recovered shapes.}}
		\label{fig_invivo_miccai}
	\end{figure*}

	\begin{figure*}[htpb]    
		\centering
		\subfloat{    
			\begin{minipage}[htpb]{0.95\textwidth}    
				\centering
				\includegraphics[width=1\linewidth]{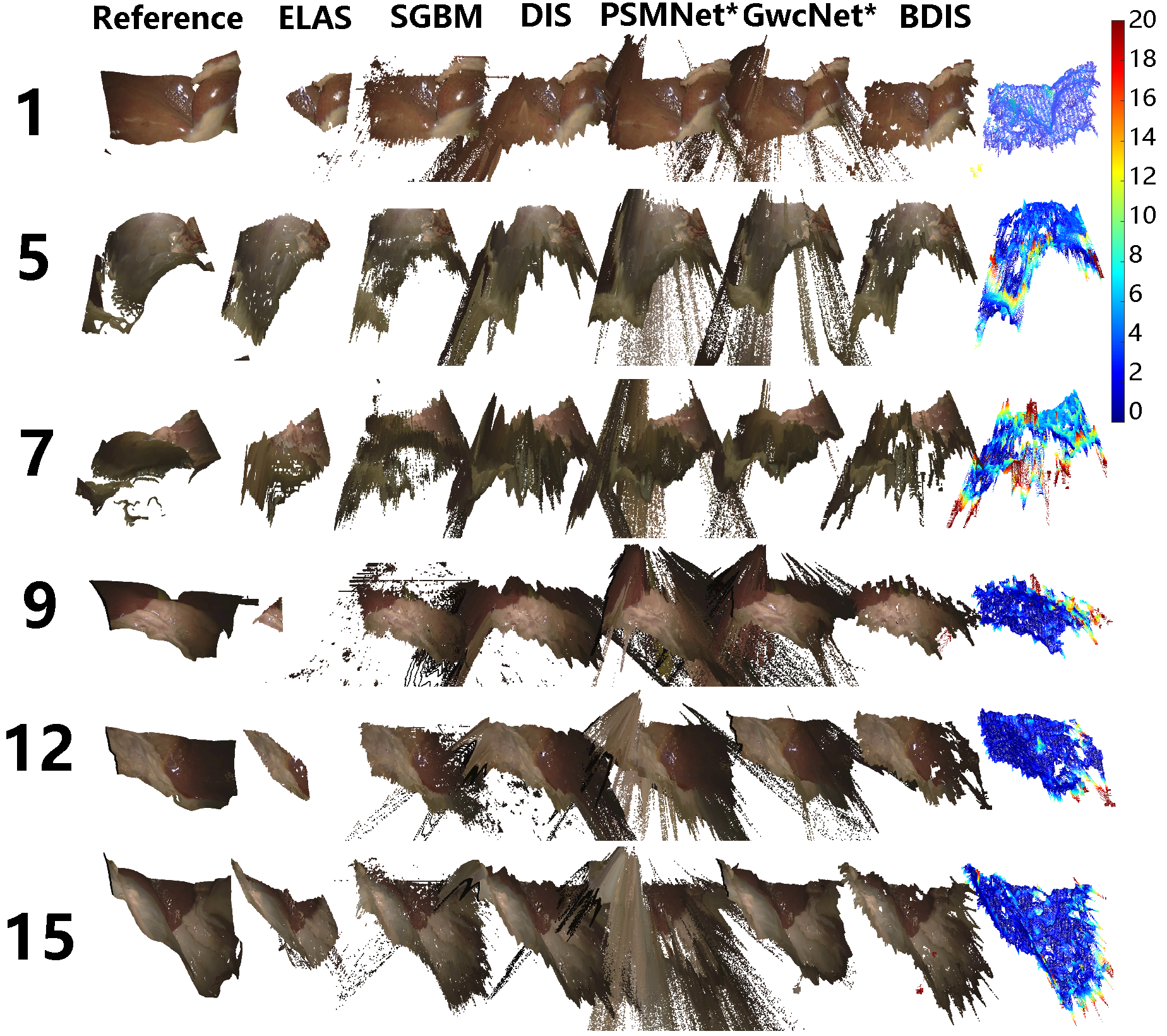}            
			\end{minipage}                
		}
		\caption{\sjw{The figure shows sample reconstructions of the SERV-CT stereo data set. The rows show data set 1, 5, 7, 9, 12 and 15 which differs significantly in texture. The error map (in mm) in the last column is the error of BDIS. Readers are encouraged to refer to the attached video for more results.}}
		\label{fig_invivo_serv}
	\end{figure*}

	All the results indicate that predictions from BDIS are similar to or slightly better than the baseline ELAS. BDIS is more efficient than other baseline methods in handling the disparity prediction in the non-Lambertian scenario. Regarding the median error, BDIS is $9.55\%$ and $24.24\%$ better than ELAS in the diffuse lighting and non-Lambertian reflectance illumination. Meanwhile, regarding the mean error, BDIS is $8.18\%$ and $6.38\%$ better than ELAS in the diffuse lighting and non-Lambertian reflectance illumination. Furthermore, BDIS has more valid predictions than ELAS, and its accuracy can be higher by filtering low probability predictions. Fig.~\ref{fig_exvivo_comparison_1} also qualitatively shows that BDIS achieves fewer ambiguities than ELAS. In general, BDIS is more robust and accurate than ELAS in both diffuse lighting and non-Lambertian reflectance illumination.\par

	\sjw{All results reveal the better performance of BDIS over DIS.} Fig.~\ref{fig_exvivo_comparison_1} and Table~\ref{Table_exvivo_dataset_colon} reveal that the bad average error comparison is caused by the small group of far-out points on the dark regions/edges. The figures and the number of valid predictions suggest BDIS produces more predictions but suffers from inaccurate dark region predictions than DIS. The small differences cannot be easily observed visually, but the quantitative results provide evidence on its side. In this article, we do not enforce any prior smoothness constraint in the fusion process. The smoothness	prior needs to be treated with caution because it impairs the accuracy in the abrupt changes on the recovered shape.\par

	The performance of DNN-based methods is much inferior in the synthetic data set test (Table~\ref{Table_exvivo_dataset_colon}) than in the following in-vivo data set. \sjw{The DNN's performances in Table~\ref{Table_exvivo_dataset_colon} deviate from the conclusions in~\cite{allan2021stereo}.} It also contradicts to the following in-vivo test (Section~\ref{section_in_vivo}). The failure should be credited to insufficient parameter tuning. Specifically, there is a large texture gap between the synthetic training data set and the data set adopted in training the pre-trained DNN model. The pre-trained model is obtained from the city-scape data set Kitti2015~\cite{geiger2013vision,Menze2018JPRS}. The texture of the synthetic data set was simulated from the game engine and bereft of many imaging details and cues. That is to say, the synthetic data set is not the nature image as the Kitti2015. Previous researches~\cite{chen2021s2r,atapour2018real,zheng2018t2net} suggest that the performance of the convolution neural network is heavily dependent on the texture of the images. This phenomenon also explains the massive effort devoted to bridging the domain gap between synthetic data and real-world data in the autonomous driving community. In our case, the pre-trained weights of kernels contribute poorly to the convergence of the model. Thus, instead of fine-tuning the pre-trained model, the DNN-based models should be trained on the synthetic data set from the beginning. Unfortunately, we could not successfully train the DNN based on the generated data sets for the two DNN-based methods. \sjw{The quantitative real-world in-vivo/ex-vivo data sets with reference are tested in Section~\ref{section_in_vivo} for further validation.}\par

	\begin{table*}[]
		\centering
		\caption{The table presents the mean absolute depth error comparisons on the SCARED stereo data set. D\textit{X}K\textit{Y} means the keyframe \textit{Y} in data set \textit{X}. The average of the mean and the average median errors of these approaches are also listed. data set 8 and 9 are designated as the testing data set in~\cite{allan2021stereo}. $\cdot^{*}$ means the method requires extra training data. Errors are in mm. The number of valid depth pixels is in $1000$. BDIS$^\triangle$ refers to our preliminary work~\cite{song2021bayesian}.}
		\begin{tabular}{p{3.11cm}<{\raggedright}|p{1.66cm}<{\centering}|p{1.66cm}<{\centering}|p{1.66cm}<{\centering}|p{1.66cm}<{\centering}|p{1.66cm}<{\centering}|p{1.66cm}<{\centering}|p{1.66cm}<{\centering}}
			\toprule 
			& ELAS                            & SGBM                            & DIS    & GwcNet$^{*}$               & PSMNet$^{*}$ & \multicolumn{1}{c|}{BDIS$^\triangle$} & BDIS                            \\ \midrule \midrule
			D8K1                 & 6.452                           & 6.522                           & 6.847  & 7.726                           & 14.759            & 6.593                     & \textbf{6.386} \\
			D8K2                 & 3.813                           & 3.785                           & 4.036  & \textbf{3.642} & 7.151             & 3.947                     & 3.913                           \\
			D8K3                 & 1.642                           & 3.258                           & 1.904  & \textbf{1.513} & 3.210             & 1.782                     & 1.754                           \\
			D8K4                 & 2.008                           & 1.912                           & 2.434  & \textbf{1.824} & 3.691             & 2.106                     & 2.105                           \\
			D8K5                 & 2.927                           & 3.187                           & 2.625  & \textbf{1.599} & 3.220             & 1.906                     & 1.880                           \\
			D9K1                 & 3.197                           & \textbf{2.539} & 3.399  & 2.670                           & 6.577             & 2.630                     & 2.607                           \\
			D9K2                 & \textbf{0.683} & 1.452                           & 0.995  & 0.826                           & 1.833             & 0.766                     & 0.704                           \\
			D9K3                 & \textbf{0.834} & 1.360                           & 1.862  & 0.859                           & 2.064             & 0.968                     & 0.940                           \\
			D9K4                 & 0.806                           & 19.094                          & 7.441  & 1.032                           & 1.220             & 0.694                     & \textbf{0.681} \\
			D9K5                 & 0.385                           & 30.608                          & 13.456 & 0.509                           & 0.888             & 0.358                     & \textbf{0.343} \\ \midrule \midrule
			Average mean error   & 2.275                           & 7.372                           & 4.500  & 2.220                           & 4.461             & 2.175                     & \textbf{2.131} \\
			Average median error & 1.825                           & 3.223                           & 3.012  & \textbf{1.556} & 3.215             & 1.820                     & 1.817                           \\
			Valid depth pixel    & 674.71                          & 560.42                          & 817.37 & 826.25                          & 826.46            & 739.06                    & 713.58                          \\ \bottomrule
		\end{tabular}
		\label{Table_MICCAI_dataset_II}
	\end{table*}
	
	\begin{table}[]
		\centering
		\caption{Table shows the mean absolute depth error comparisons on the SCARED stereo data set. D\textit{X}K\textit{Y} is the keyframe indexed as \textit{Y} in data set \textit{X}. Results of the prior-free methods ELAS, SGBM, DIS, and the proposed BDIS are presented. The cells are the average absolute mean error (in mm). BDIS$^\triangle$ refers to our preliminary work~\cite{song2021bayesian}.}
		\begin{tabular}{p{1.14cm}<{\raggedright}|p{1.04cm}<{\centering}|p{1.04cm}<{\centering}|p{1.04cm}<{\centering}|p{1.04cm}<{\centering}|p{1.04cm}<{\centering}}
			\toprule
			& ELAS                            & SGBM                            & DIS    & BDIS$^\triangle$    & BDIS                            \\ \midrule \midrule
			D1K1                 & \textbf{0.673} & 0.959                           & 1.350  & 0.754   & 0.749                           \\
			D1K2                 & \textbf{0.981} & 1.162                           & 2.042  & 1.099   & 1.075                           \\
			D1K3                 & 0.905                           & \textbf{0.844} & 1.617  & 0.958   & 0.945                           \\
			D1K5                 & \textbf{0.794} & 2.822                           & 1.735  & 0.822   & 0.812                           \\
			D2K1                 & 0.718                           & 8.080                           & 2.590  & 0.702   & \textbf{0.690} \\
			D2K2                 & 1.422                           & \textbf{1.112} & 3.281  & 1.142   & 1.143                           \\
			D2K3                 & \textbf{1.071} & 4.115                           & 8.105  & 1.099   & 1.086                           \\
			D2K4                 & 0.367                           & 43.066                          & 4.734  & 0.352   & \textbf{0.345} \\
			D2K5                 & 1.624                           & 1.888                           & 2.648  & 1.496   & \textbf{1.483} \\
			D3K1                 & 2.147                           & \textbf{1.522} & 2.477  & 1.688   & 1.673                           \\
			D3K2                 & 1.932                           & 4.728                           & 1.513  & 1.050   & \textbf{1.033} \\
			D3K3                 & 2.113                           & 1.911                           & 2.739  & 1.568   & \textbf{1.549} \\
			D3K4                 & 2.394                           & 5.404                           & 2.272  & 1.040   & \textbf{1.017} \\
			D3K5                 & 2.006                           & 1.558                           & 4.101  & 1.539   & \textbf{1.517} \\
			D6K1                 & \textbf{4.084} & 5.216                           & 7.343  & 5.231   & 5.238                           \\
			D6K2                 & \textbf{2.613} & 2.967                           & 4.186  & 2.879   & 2.877                           \\
			D6K3                 & \textbf{2.575} & 3.293                           & 4.264  & 3.096   & 3.065                           \\
			D6K4                 & 4.719                           & 4.703                           & 6.807  & 4.562   & \textbf{4.554} \\
			D6K5                 & 8.842                           & 6.261                           & 7.878  & 6.503   & \textbf{6.162} \\
			D7K1                 & \textbf{3.685} & 6.153                           & 4.748  & 4.609   & 4.583                           \\
			D7K2                 & 4.194                           & 4.203                           & 5.931  & 3.296   & \textbf{3.285} \\
			D7K3                 & \textbf{1.256} & 11.685                          & 2.709  & 1.406   & 1.345                           \\
			D7K4                 & \textbf{2.171} & 4.914                           & 2.831  & 2.664   & 2.551                           \\
			D7K5                 & \textbf{3.296} & 3.555                           & 3.954  & 3.592   & 3.301                           \\
			D8K1                 & 6.452                           & 6.522                           & 6.847  & 6.498   & \textbf{6.386} \\
			D8K2                 & 3.813                           & \textbf{3.785} & 4.036  & 3.902   & 3.913                           \\
			D8K3                 & \textbf{1.642} & 3.258                           & 1.904  & 1.756   & 1.754                           \\
			D8K4                 & 2.008                           & \textbf{1.912} & 2.434  & 2.186   & 2.105                           \\
			D8K5                 & 2.927                           & 3.187                           & 2.625  & 1.969   & \textbf{1.880} \\
			D9K1                 & 3.197                           & \textbf{2.539} & 3.399  & 2.998   & 2.607                           \\
			D9K2                 & 0.683                           & 1.452                           & 0.995  & \textbf{0.680} & 0.704                           \\
			D9K3                 & \textbf{0.834} & 1.360                           & 1.862  & 0.968   & 0.940                           \\
			D9K4                 & 0.806                           & 19.094                          & 7.441  & 0.689   & \textbf{0.681} \\
			D9K5                 & 0.385                           & 30.608                          & 13.456 & 0.349   & \textbf{0.343} \\ \midrule \midrule
			Mean   & \textbf{1.6462}                   & 4.196                          & 1.913 & 1.807  & 1.778                          \\
			Median & 2.333                        & 6.054                           & 4.025 & 2.219  & \textbf{2.158}                         \\
			Validity    & 674.73                           & 549.61                           & 803.85  & 739.00   & 717.25                           \\ \bottomrule 
		\end{tabular}
		\label{Table_MICCAI_dataset_I_individual}
	\end{table}

	\begin{table}[]
	\setlength\tabcolsep{5.5pt} 
		\centering
		\caption{\sjw{Table shows the mean absolute depth error comparisons on the SERV-CT stereo data set. Left column is the keyframe indexed. Results of the prior-free methods ELAS, SGBM, DIS, GwcNet, PSMNet and the proposed BDIS are presented. The cells are the average absolute median error (in mm).}}
		\sjw{
		\begin{tabular}{p{0.95cm}<{\raggedright}|p{0.75cm}<{\centering}|p{0.85cm}<{\centering}|p{0.75cm}<{\centering}|p{0.99cm}<{\centering}|p{1.03cm}<{\centering}|p{0.7cm}<{\centering}}
        \toprule
         & ELAS    & SGBM    & DIS     & GwcNet  & PSMNet  & BDIS    \\ \midrule \midrule
1        & 1.027  & 2.622  & 3.442  & 3.227  & 5.295  & 3.315  \\
2        & 2.976  & 4.011  & 4.261  & 4.629  & 6.919  & 4.024  \\
3        & 1.641  & 4.390  & 3.226  & 3.343  & 4.655  & 2.977  \\
4        & 2.485  & 5.202  & 2.854  & 3.967  & 4.149  & 2.538  \\
5        & 5.344  & 5.674  & 5.115  & 5.693  & 9.653  & 5.122  \\
6        & 2.091  & 3.883  & 3.616  & 3.471  & 5.550  & 3.554  \\
7        & 2.004  & 6.177  & 5.806  & 6.238  & 8.481  & 5.608  \\
8        & 1.633  & 3.491  & 3.424  & 6.981  & 6.097  & 3.295  \\
9        & 4.242  & 1.986  & 2.593  & 5.539  & 3.165  & 1.632  \\
10       & 2.262  & 1.870  & 2.092  & 4.057  & 3.516  & 1.640   \\
11       & 1.596  & 1.681  & 2.068  & 4.467  & 2.516  & 1.677  \\
12       & 1.595  & 1.399  & 1.476  & 2.763  & 2.405  & 1.235  \\
13       & 6.718  & 3.772  & 2.781  & 3.672  & 2.687  & 1.771  \\
14       & 9.307  & 16.437 & 3.367  & 10.848 & 2.437  & 0.595 \\
15       & 3.589  & 1.768  & 1.697  & 6.546  & 2.480  & 1.590  \\ \midrule \midrule
Mean     & 5.665  & 5.478  & 7.602  & 12.750 & 8.585  & 3.750   \\
Median   & 3.233  & 4.291  & 3.187  & 5.029  & 4.667  & 2.705   \\
Validity & 120.00 & 184.83 & 398.30 & 379.56 & 412.02 & 311.48 \\  \bottomrule 
\end{tabular}
}
		\label{Table_SERV_CT_individual}
	\end{table}

	\begin{table*}[]
		\caption{\sjw{The table presents the quantitative comparisons between the DIS, SGBM and BDIS regarding the shape of ELAS as the reference.} The results are presented as the average absolute depth error (mm). }
		\label{Table_overall_accuracycompare}
		\centering
		\begin{tabular}{p{1.39cm}<{\raggedright}|p{1.39cm}<{\centering}|p{1.39cm}<{\centering}|p{1.39cm}<{\centering}|p{1.39cm}<{\centering}|p{1.39cm}<{\centering}|p{1.39cm}<{\centering}|p{1.39cm}<{\centering}|p{1.39cm}<{\centering}|p{1.39cm}<{\centering}}
			\toprule 
			& \multicolumn{3}{c|}{Median error}                & \multicolumn{3}{c|}{Average error}               & \multicolumn{3}{c}{Standard deviation}          \\ \midrule \midrule
			& DIS  & \multicolumn{1}{c|}{SGBM} & BDIS          & DIS  & \multicolumn{1}{c|}{SGBM} & BDIS          & DIS  & \multicolumn{1}{c|}{SGBM} & BDIS          \\ \midrule \midrule
			Data 1 & 1.81 & \textbf{1.29}             & 1.64          & 2.69 & \textbf{1.96}             & 2.30          & 2.83 & \textbf{2.15}             & 2.25          \\
			Data 2 & 1.06 & 1.78                      & \textbf{0.91} & 1.80 & 2.79                      & \textbf{1.37} & 2.45 & 3.03                      & \textbf{1.67} \\
			Data 3 & 0.46 & 0.44                      & \textbf{0.41} & 0.87 & 0.67             & \textbf{0.65} & 1.34 & 0.90                      & \textbf{0.89} \\
			Data 4 & 0.49 & 0.41                      & \textbf{0.39} & 1.38 & \textbf{0.76}             & 1.06          & 2.24 & \textbf{1.02}             & 1.90          \\
			Data 5 & 0.52 & 0.55                      & \textbf{0.42} & 1.02 & 0.93                      & \textbf{0.80} & 1.62 & \textbf{1.28}             & \textbf{1.28} \\ \bottomrule
		\end{tabular}
	\end{table*}

	\subsection{Quantitative Comparisons on the in-vivo/ex-vivo data set}
	\label{section_in_vivo}
	
	\sjw{On top of the qualitative comparison, we validated the performance of BDIS on the real-world SCARED (in-vivo) and SERV-CT (ex-vivo) stereo data sets. For the SCARED data set, two separate experiments were conducted.} In the first experiment, following the recommendation of~\citet{allan2021stereo}, only the last ten image pairs (images in data sets 8 and 9) were tested since the DNN-based methods require the rest images (24 stereo image pairs) for fine-tuning the network. Among the 24 stereo image pairs, 18 were adopted to fine-tune PSMNet and GwcNet, while the validation was done on the other 6 stereo image pairs. The fine-tuned networks were further applied to predict the rest 10 testing data sets. Fig.~\ref{fig_invivo_miccai} shows sample reconstructions of the baseline methods. In the second experiment for prior-free methods only, since the prior-free approaches are free of training data, all available (34 totally) stereo image pairs were used for more comprehensive comparisons. \par

	The frame-wise mean absolute depth errors between the state-of-the-art approaches and the proposed BDIS on the SCARED stereo data set are demonstrated in Table~\ref{Table_MICCAI_dataset_II}. The last 10 stereo image pairs of the testing keyframes designated by~\citet{allan2021stereo} are individually presented. It also summarizes the median error, average error, and the number of valid depth pixel predictions of these approaches. The results indicate that BDIS is $6.33\%$ more accurate than ELAS regarding the average median error. Compared with ELAS, the proposed BDIS improves the performance on the average median error by $0.44\%$. Nevertheless, BDIS has $5.67\%$ more valid predictions over ELAS. Moreover, Table~\ref{Table_MICCAI_dataset_I_individual} indicates that SGBM can predict similar depth as BDIS and ELAS, but it is not robust. It shows that the trained GwcNet is comparable to BDIS and ELAS. The results of GwcNet validate our assumption (Section~\ref{section_synthetic}) and conclusions from the researches~\cite{chen2021s2r,atapour2018real,zheng2018t2net} that the performance of DNN is dependent on the texture similarity between the training and testing images. The images in the SCARED stereo data set are natural images, and their textures are much more abundant than the synthetic data set (results in Section~\ref{section_synthetic}). However, it should be emphasized that the accuracy of the DNN-based method is heavily dependent on the quality of the training data set. The comparisons of the prior-free and the DNN-based methods are for reference only. In general, it confirms our claim that BDIS and ELAS have similar accuracy and predicts the same amount of valid depth.\par

	Table~\ref{Table_MICCAI_dataset_I_individual} shows the comparisons of the prior-free methods ELAS, SGBM, DIS, and BDIS. The frame-wise mean and median errors of all keyframes are summarized. In general, the accuracy of ELAS is $7.42\%$ higher in average mean error over BDIS, while BDIS is $7.55\%$ better regarding the average median error. Moreover, the valid depth pixels of BDIS are $5.93\%$ more than ELAS. This result is consistent with other quantitative experiments in this article. \par
	
	\sjw{ELAS, SGBM, DIS, PSMNet, GwcNet, and BDIS were also tested on the SERV-CT data set. Since only 16 image pairs are provided in SERV-CT, the trained DNNs of PSMNet and GwcNet on SCARED data sets were adopted directly. Thus, the results of PSMNet and GwcNet are cross the textural domains. Sample qualitative comparisons are presented in Fig.~\ref{fig_invivo_serv}. Quantitative comparison is shown in Table~\ref{Table_SERV_CT_individual}. Please note that only first 15 data sets were tested because ELAS did not yield a valid prediction of the $16$th image pair.}\par
	
	\sjw{Fig.~\ref{fig_invivo_serv} and Table~\ref{Table_SERV_CT_individual} indicate that ELAS has much smaller valid predictions compared to BDIS, which contradicts the results in SCARED experiments. The reason is that the region of valid pixels is within the Delaunay polygon defined by the sparse supporting corner points. Since the textural and illuminational quality of images in SERV-CT (Fig.~\ref{fig_SERV_CT_samples}) data are worse than images in SCARED (Fig.~\ref{fig_MICCAI_samples}), the sparse corner points matching module in ELAS does not provide satisfying point pairs. Thus, the accuracy of ELAS with small valid regions is for reference only. In general, BDIS achieves the best mean and median absolute error performance than the other three prior-free methods.}\par
	
	\sjw{Table~\ref{Table_SERV_CT_individual} also shows that the two DNN methods are not satisfying. The major reason is that both DNN methods predicted many outliers on the edges, which suffer from the darkness. Thus, their average median errors are much better than the mean error. Another reason (minor) may be the cross-domain application because the image textures of SERV-CT are different from SCARED. Moreover, although the table shows that DNNs' results are not satisfying, it should be emphasized that  Fig.~\ref{fig_invivo_serv} shows that the central region of the two DNN methods are acceptable.} \par

	\begin{figure*}[htpb]    
\centering
\captionsetup[subfigure]{labelformat=empty}
\begin{minipage}[]{1\textwidth}
\centering
\includegraphics[width=1\linewidth]{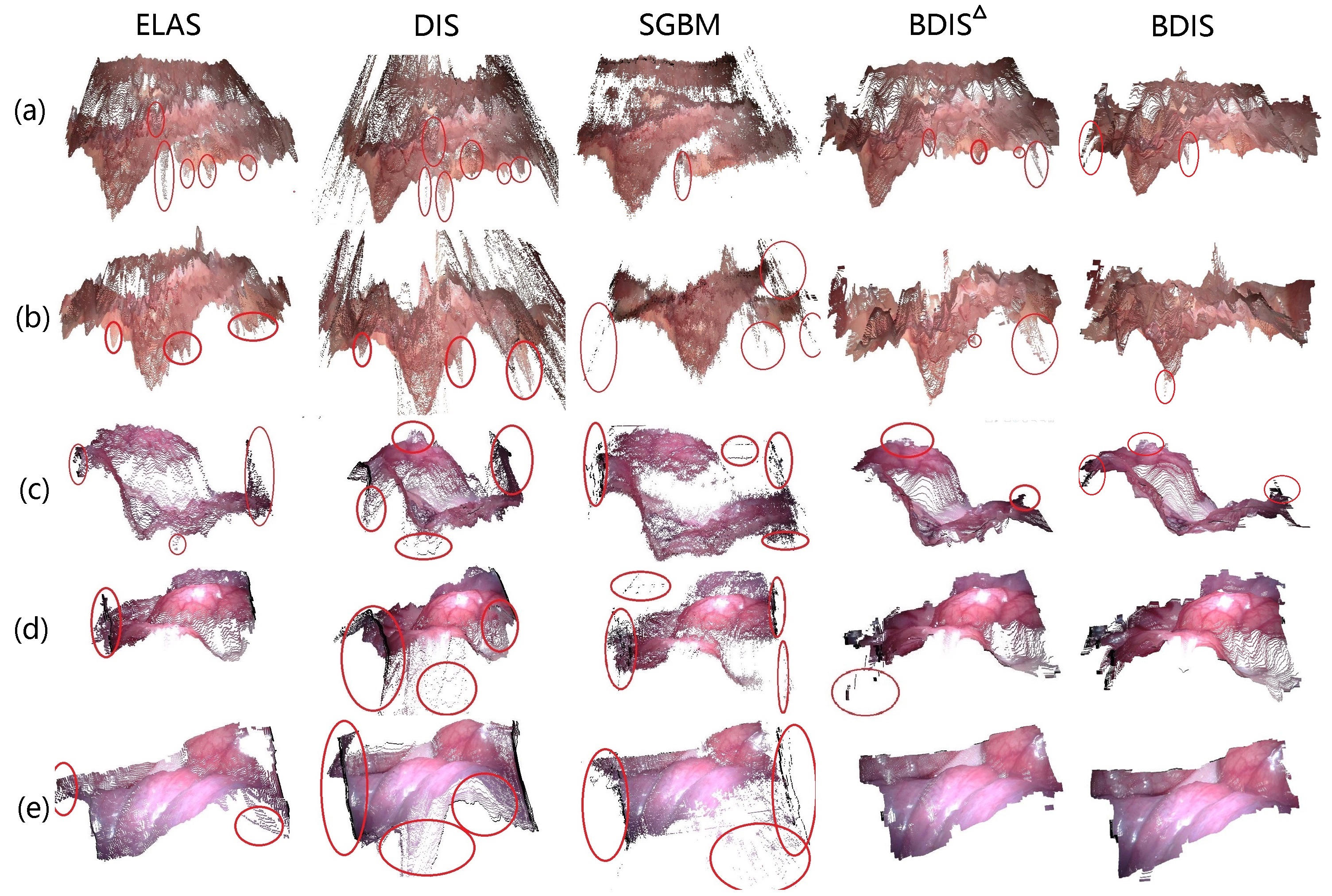}
\end{minipage}
\caption{\sjw{The figure shows sample recovered models of the 5 classes. BDIS$^\triangle$ refers to the BDIS proposed in~\cite{song2021bayesian}.}}
\label{fig_quantitative_compare}
\end{figure*}

\begin{figure*}[htpb]    
\centering
\begin{minipage}[]{1\textwidth}
\centering
\includegraphics[width=1\linewidth]{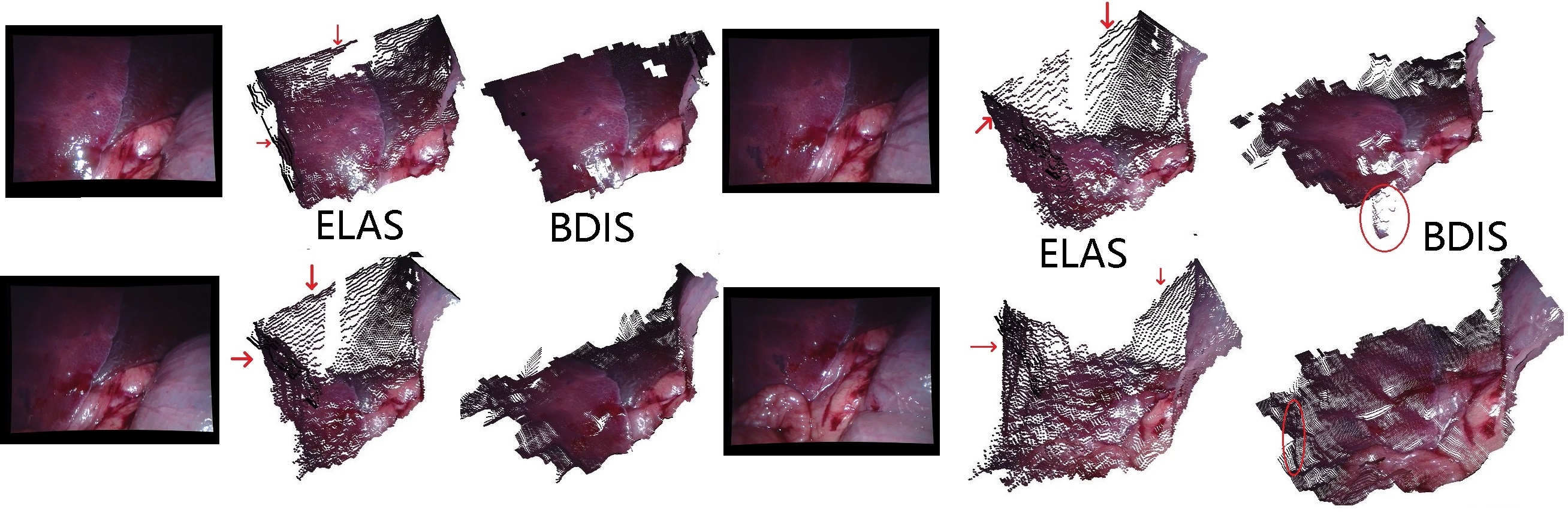}
\end{minipage}
\caption{The figure illustrates the qualitative comparisons of ELAS~\cite{geiger2010efficient}, DIS~\cite{kroeger2016fast} and the proposed method. The marked regions are the wrongly reconstructed.}
\label{fig_quantitative_ELAS}
\end{figure*}

	\subsection{Qualitative comparison with ELAS on in-vivo data set}
	\label{section_in_vivo_ELAS}
	
	The proposed BDIS was compared with ELAS, DIS, and SGBM on the in-vivo Hamlyn data sets. \sjw{Only qualitative comparisons of the prior-free methods can be given on this data set because no reference labels are provided for training DNNs.} Like the SCARED data set, the Hamlyn data set was collected from the stomach of the porcine, which provides more abundant textured stereo images. This data set was employed to show the qualitative comparisons of the prior-free approach and the quantitative difference between the predictions of ELAS and BDIS. The experiment validates our claim that the proposed BDIS has similar performance to the baseline ELAS in the scenario of abundant textures.\par

	The Hamlyn data set is categorized into $5$ groups by the camera-to-surface distance since the distances in this data set vary significantly. Fig.~\ref{fig_quantitative_compare} shows the sample comparisons between baseline prior-free methods and BDIS. Table~\ref{Table_overall_accuracycompare} shows the average accuracy of all frames within each category. It indicates that the proposed method achieves around $10\%$ (median error) and $15\%$ accuracy over the original DIS algorithm. \sjw{Although no reference is provided, this comparison shows BDIS achieves an average $0.4 - 1.66 mm$ (median error) and $0.65 - 2.32 mm$ (mean error) difference from ELAS's results.}\par

	Fig.~\ref{fig_quantitative_compare} shows BDIS can achieve similarly or even slightly better robustness in handling the photometric inconsistency in the stereo matching process. The photometric inconsistency poses great difficulty to the surgical stereo matching process since the stereo images are vulnerable to non-Lambertian reflectance, dark region, or textureless surfaces. Fig.~\ref{fig_invivo_miccai} and Fig.~\ref{fig_quantitative_compare} indicate BDIS has fewer outliers at the image edges. BDIS overcomes this issue mainly due to three strategies: probabilistic inverse residual-based patch fusion, initialization strategy, and coarse-to-fine probability propagation. The probabilistic inverse residual-based patch fusion uses the predictions from multiple patches to mitigate the invalid disparity pixels (ambiguous predictions). These invalid predictions mainly result from the edges of the rectified image after the image undistortion, occlusions from objects to the left or right cameras, or unsuccessful predictions due to insufficient information. The dubious and unsuccessful predictions in coarse level pass incorrect initialization to the finer scale. Our probabilistic inverse residual-based patch fusion quantifies the posterior probability; the initialization strategy discards the patch that does not converge; coarse-to-fine probability propagation provides an important indicator to help lower the patches' probabilities with invalid pixels. Since there are enough overlaps between the neighboring patches, the low confidence initialization may be compensated by its neighbors. In the worst case, following ELAS, the predictions with very low probability are removed.\par
	
	On top of the initialization issue, the ambiguous local minima pose noticeable difficulty in the stereo matching process. The local minima issue has a heavy impact on the performance of fast LK since the corresponding cost function under the photometric consistency assumption is non-convex. The toy example in Fig.~\ref{fig_prob_density} shows that our estimated coarse-to-fine strategy measures the probability well, and the fusion module alleviates the ambiguity of the disparity with the neighboring patch predictions. This article's sample figures demonstrate that BDIS has fewer local minima than the baseline prior-free approaches. This is also one reason for the high accuracy of BDIS in the qualitative comparisons. However, it should be addressed that we do not enforce any prior smoothness constraint in the optimization process. The smoothness prior needs to be treated with caution because it impairs the accuracy of the abrupt changes and edges on the recovered shape.

	In addition to testing the well-illuminated soft tissue shown in Fig.~\ref{fig_prob_density}, we further presented the scenario of bad illumination, which corresponds to the non-Lambertian data set (Fig.~\ref{fig_exvivo_comparison_1}). In these cases, the light intensity is roughly (not accurate) proportional to $\operatorname{cos}(\alpha)$ where $\alpha$ is the angle between the surface's normal and the viewing direction. The texture vanishes when $\alpha$ is close to zero (sharply lighted patch) or when $\alpha$ is large (dark region). Both bad-textured regions are obstacles to the cost function. As demonstrated in Fig.~\ref{fig_quantitative_ELAS}, the center of the soft tissue is exposed to intense lighting while the marginal region is dark. Note that the marginal region is dark but not invalid. As depicted in Fig.~\ref{fig_quantitative_ELAS}, our proposed BDIS outperforms ELAS in estimating or filtering dark pixels. \par

	\subsection{Comments on the comparison with DNN-based methods}
	
	\sjw{We provide our comments on the comparisons with DNN-based methods in Section~\ref{section_synthetic} and \ref{section_in_vivo}.} It should be emphasized that the comparison with the DNN-based methods is for the completeness of all the comparisons, and our presented conclusions regarding DNN-based methods are for reference only. First, comparing prior-free methods with DNN-based methods is not fair. DNN-based methods' performance heavily relies on the size and quality of the training data, which varies greatly in CAS. Up to now, learning-based methods cannot fully substitute prior-free methods due to prior-free methods' robustness to parameter tuning and being free of training data. Thus, the conclusion that DNN-based methods work in one case cannot be generalized to other cases. Moreover, DNN consumes heavy computational resources of GPU, while all the prior-free methods use only one CPU core. Additionally, BDIS and DNN methods can be complementary. Since DNN is on GPU and dependent on the training data set, CPU-based prior-free BDIS can assist in data fusion or cross-validation. \par



	\begin{table}[]
		\centering
		\caption{\sjw{The table shows the coverage rate of the estimated pixel-wise variance~\eqref{Eq_var_depth}. It summarizes the ratio of pixel-wise errors that falls within the bound $\mathrm{e}_{\mathbf{x}}$ falls within the range of $\left[- 1.96
			\hat{\sigma}_{x}, 1.96\hat{\sigma}_{x} \right]$. The ideal coverage rate is $95\%$. A and a are the results on the synthetic data sets colon in diffuse lighting and non-Lambertian reflectance. S\textit{X} are the results on SERV-CT data set. The results of the in-vivo data sets are also tested. It is marked as D\textit{X}K\textit{Y} where \textit{Y} is the keyframe index and is \textit{X} is data set index. All values are in percentage.}}
		\sjw{
		\begin{tabular}{p{0.69cm}<{\raggedright}p{0.69cm}<{\raggedright}p{0.69cm}<{\raggedright}p{0.69cm}<{\raggedright}p{0.69cm}<{\raggedright}p{0.69cm}<{\raggedright}p{0.69cm}<{\raggedright}p{0.69cm}<{\raggedright}}
A1    & A2                        & A3                        & A4    & A5    & A6    & A7    & A8                   \\ \hline
93.87 & 92.35                     & 87.37                     & 64.01 & 90.13 & 65.76 & 82.35 & 89.02                \\ \hline
A9    & A10                       & a1                        & a2    & a3    & a4    & a5    & a6                   \\ \hline
88.48 & 92.28                     & 93.09                     & 92.11 & 86.72 & 56.9  & 90.9  & 90.75                \\ \hline
a7    & a8                        & a9                        & a10   & S1    & S2    & S3    & S4                   \\ \hline
85.33 & 85.66                     & 73.26                     & 93.52 & 46.27 & 42.16 & 34.56 & 57.30                \\ \hline
S5    & S6    & S7    & S8    & S9    & S10   & S11   & S12                  \\ \hline
67.30 & 49.25 & 64.04 & 64.32 & 93.56 & 93.68 & 94.00 & 92.71                \\ \hline
S13   & S14                       & S15  & D1K1  & D1K2  & D1K3  & D1K5  & D2K1                 \\ \hline
90.25 & 93.35                     & 94.46 & 93.04 & 92.3  & 92.81 & 91.29 & 87.01                \\ \hline
D2K2  & D2K3                      & D2K4                      & D2K5  & D3K1  & D3K2  & D3K3  & D3K4                 \\ \hline
97.06 & 84.53                     & 87.43                     & 90.94 & 93.82 & 89.21 & 89.79 & 90.35                \\ \hline
D3K5  & D6K1                      & D6K2                      & D6K3  & D6K4  & D6K5  & D7K1  & D7K2                 \\ \hline
90.49 & 66.38                     & 45.02                     & 57.98 & 57.8  & 50.64 & 50.87 & 63.51                \\ \hline
D7K3  & D7K4                      & D7K5                      & D8K1  & D8K2  & D8K3  & D8K4  & D8K5                 \\ \hline
47.97 & 38.66                     & 35.01                     & 50.35 & 33.12 & 52.61 & 43.92 & 61.84                \\ \hline
D9K1  & D9K2                      & D9K3                      & D9K4  & D9K5  &       &       & \multicolumn{1}{l}{} \\ \hline
80.75 & 88.36                     & 85.68                     & 74.13 & 83.77 &       &       & \multicolumn{1}{l}{} \\ \hline
\end{tabular}
		}
		\label{Table_var}
	\end{table}
	
	\begin{table*}[]
		\centering
		\caption{The table shows the average mean and the median absolute depth error comparisons on the SCARED stereo data set and the synthetic data set colon (in diffuse lighting and non-Lambertian reflectance). The full BDIS, BDIS without GMM, BDIS without a coarse-to-fine strategy, and BDIS without dynamic sigma $\sigma_r^{(n)}$ ($\sigma_r^{(n)}$ is fixed to $8$ in our preliminary research~\cite{song2021bayesian}) are presented. For the benefits of analyzing, the normalized results are also provided.}
		\sjw{
		\begin{tabular}{p{2.34cm}<{\raggedright}|p{3.04cm}<{\raggedright}|p{2.44cm}<{\centering}|p{2.44cm}<{\centering}|p{2.64cm}<{\centering}|p{2.64cm}<{\centering}}
			\toprule
			&                    & BDIS  & \begin{tabular}[c]{@{}c@{}}BDIS \\ (no sGMM)\end{tabular} & \begin{tabular}[c]{@{}c@{}}BDIS\\  (no Coarse-to-fine)\end{tabular} & \begin{tabular}[c]{@{}c@{}}BDIS \\ (no dynamic sigma)\end{tabular} \\ 
			\midrule \midrule
			\multirow{4}{*}{SCARED}                                                                & Median             & 1.778 & 1.792                                                    & 1.799                                                               & 1.784                                                              \\
			& Mean               & 2.158 & 2.204                                                    & 2.197                                                               & 2.213                                                              \\
			& Median (Normalize) & 1.000 & 1.008                                                    & 1.012                                                               & 1.036                                                              \\
			& Mean (Normalize)   & 1.000 & 1.021                                                    & 1.018                                                               & 1.025                                                              \\ 
			\midrule \midrule
			\multirow{4}{*}{\begin{tabular}[l]{@{}l@{}}Synthetic \\ (diffuse lighting)\end{tabular}} & Median             & 0.161 & 0.160                                                    & 0.163                                                               & 0.163                                                              \\
			& Mean               & 0.202 & 0.210                                                    & 0.207                                                                 & 0.207                                                              \\
			& Median (Normalize) & 1.000 & 0.994                                                    & 1.012                                                               & 1.012                                                              \\
			& Mean (Normalize)   & 1.000 & 1.040                                                    & 1.020                                                               & 1.020                                                              \\ 
			\midrule \midrule
			\multirow{4}{*}{\begin{tabular}[l]{@{}l@{}}Synthetic \\ (non-Lambertian)\end{tabular}}   & Median             & 0.138 & 0.140                                                    & 0.138                                                               & 0.139                                                              \\
			& Mean               & 0.224 & 0.247                                                    & 0.240                                                                 & 0.236                                                             \\
			& Median (Normalize) & 1.000 & 1.020                                                    & 1.000                                                               & 1.007                                                              \\
			& Mean (Normalize)   & 1.000 & 1.102                                                    & 1.071                                                               & 1.053                                                              \\ \bottomrule
		\end{tabular}
		}
		\label{Table_ablation_study}
	\end{table*}

	\subsection{Time consumption}
	
	Table~\ref{Table_time_consumption} presents the time consumption comparisons in predicting the disparity of the prior-based methods ELAS, DIS, SGBM, BDIS, and the prior-based methods PSMNet and GwcNet. The experiments of the prior-free methods are all carried out on a single core of the CPU (i5-9400). The experiments of prior-based methods are implemented on GPU (GTX 1080ti). The processing time of BDIS is double as DIS owing to the extra time spent on patch-wise window traversing. Since the size of the window $\mathcal{P}'$ is chosen as $5$, BDIS needs to spend quintuple time on the patch convolution. In general, BDIS achieves similar/better performance over ELAS and runs 4-5 times faster. Table~\ref{Table_time_consumption} shows BDIS achieves the second-fastest speed (except DIS) over the rest algorithms with $17 Hz$ and $14 Hz$ respectively on both datasets. The frame rate meets the requirement for real-time SLAM~\cite{song2017dynamic,song2018mis} and AR~\cite{haouchine2013image,widya20193d} tasks in CAS. The time consumption comparison validates our claim that BDIS is the only real-time single-core CPU-level stereo matching in CAS. BDIS can substitute near real-time ELAS because it is much faster and similar or slightly better regarding accuracy.\par
	
	\sjw{Moreover, readers may notice that the proposed BDIS is almost two times faster than our preliminary work~\cite{song2021bayesian}.} The reason is that we reduce the patch ratio from $0.6$ to $0.55$. A smaller patch ratio means much fewer patches to be processed. Our new strategies, coarse-to-fine and dynamic sigma $\sigma_r^{(n)}$, contribute to better accuracy, and thus we can sacrifice small accuracy in exchange for faster speed. The new strategies consume very limited extra time in the implementation. Considering this, we balanced the time speed and performance by reducing $0.6$ to $0.55$.\par

	\begin{table}[]
		\centering
		\caption{\sjw{llustrated is the average of 10 times of time consumption of algorithms PSMNet, GwcNet, ELAS, DIS, SGBM, our preliminary work~\cite{song2021bayesian} and the proposed method. The two prior-based methods are marked as $^*$ because the computation is conducted on GPU. BDIS$^\triangle$ refers to our preliminary work~\cite{song2021bayesian}. BDIS(VAR) refers to enabling the optional variance estimation function. The time consumption is measured in seconds.}}
		\sjw{
		\begin{tabular}{p{1.64cm}<{\raggedright}|p{1.84cm}<{\centering}|p{1.84cm}<{\centering}|p{1.84cm}<{\centering}}
			\toprule
			& Synthetic $640 \times 480$ & SCARED $1280 \times 720$ & SERV-CT $720 \times 576$ \\ \midrule \midrule
			PSMNet$^*$ & 0.327     & 0.566   & 0.431  \\
			GwcNet$^*$ & 0.284     & 0.430   & 0.391  \\
			ELAS   & 0.246     & 0.291  & 0.264  \\
			SGBM   & 0.124     & 0.346  & 0.219\\
			DIS    & 0.038     & 0.093  & 0.054\\
			BDIS$^\triangle$   & 0.097  & 0.131  & 0.114 \\
			BDIS(VAR)   & 0.072     & 0.104  & 0.087\\
			BDIS   & 0.064     & 0.072  & 0.068\\ \bottomrule
		\end{tabular}
		}
		\label{Table_time_consumption}
	\end{table}

	\subsection{Variance estimation}
	
	This section validates the optional variance estimation technique described in Section~\ref{section_var}. \sjw{Ideally, $95\%$ of absolute depth error $\mathrm{e}_{\mathbf{x}}$ falls within the range of $\left[- 1.96
	\hat{\sigma}_{x}, 1.96\hat{\sigma}_{x} \right]$.}\par
	
	Table~\ref{Table_var} shows the frame-wise coverage rate. The experiments indicate that the variance estimation algorithm performs unstably. It indicates that results with higher accuracy have more accurate variance estimation. As exemplified in the accuracy of in-vivo experiments (Table~\ref{Table_MICCAI_dataset_I_individual}), the average mean error less than $2$ mm (data set 1, 2, 3, and 9) has a much better coverage rate. The trend in the test of the synthetic data set is not that apparent as in the in-vivo case. \sjw{One explanation is that the disparity largely deviates from the reference, and the local fine-scale variance fails to describe the uncertainty.} The large errors in the coverage rate of some frames confirm our claim that the variance estimation is partial and inaccurate. The aleatoric uncertainty can be heavily affected by the epistemic uncertainty in the fast LK process. Moreover, a large number of pixels do not obey the Gaussian probability presumption. Finally, table~\ref{Table_time_consumption} shows an extra $12\%$ to $40\%$ percent of processing time of in variance quantification.

	\subsection{Ablation study}
	
	To investigate the contributions of the modules in the proposed BDIS, several ablation studies were conducted.  \sjw{We tested the performance of the sGMM (Section~\ref{section_2_3}), coarse-to-fine probability propagation (Section~\ref{section_2_4}) and the dynamic choice of $\sigma_r^{(n)}$ (Section~\ref{section_2_2}) on the SCARED stereo data set and the synthetic data set (in diffuse lighting and non-Lambertian reflectance).} CRFs (Section~\ref{section_2_1}) was not tested in the ablation study because it is the major framework.\par 

	Table~\ref{Table_ablation_study} summarizes the accuracy (average mean and average median) comparisons. \sjw{The sGMM module was first disabled by setting the pixels in the patch in equal probabilities.} We only observe that BDIS is inferior in the average median error in the synthetic diffuse lighting data set. Next, the coarse-to-fine strategy was disabled, and the obtained probability only indicates the processing level. All results suggest the accuracy is improved from $1-3\%$. The coarse-to-fine strategy is the major innovation regarding the preliminary work~\cite{song2021bayesian}. Table~\ref{Table_exvivo_dataset_colon}~\ref{Table_MICCAI_dataset_II}~\ref{Table_MICCAI_dataset_I_individual} all validate its performance. Lastly, the dynamic sigma $\sigma_r^{(n)}$ enables tuning-free parameters and slightly better accuracy. Finally, our experiments show that all the new strategies in this paper have negligible impact on time consumption.\par

	\section{Conclusion}
	\label{section_conclusion}
	
	This article proposes BDIS as the first CPU-level real-time prior-free stereo matching in the surgical scenario. It is both novel and practical in many real-world CAS applications, e.g., lacking (high-end) GPUs, saving GPUs for other tasks, lacking annotated surgical data set for training DNN model, etc. The proposed BDIS inherits the speed of the fast LK algorithm and overcomes major obstacles in surgical image stereo matching, i.e., textureless/dark/non-Lambertian reflectance tissue surfaces, \sjw{by adopting three strategies to the deterministic fast LK: sGMM, CRFs, and a Bayesian coarse-to-fine probability propagation techniques.} The proposed BDIS correctly describes the relative confidence of the pixel-wise disparity. A variance estimation algorithm is also introduced based on the estimated probability. In this way, the pixels with low confidence are filtered out. \par

	Experiments show that the prior-free BDIS achieves an average $17$ Hz on $640 \times 480$ image with a single core of the CPU (i5-9400) for surgical images, which satisfies most SLAM, AR, and VR requirements. Its accuracy is also slightly better than the popular near real-time ELAS. The proposed method correctly quantifies the pixel-wise relative probability, which benefits outlier filtering steps. The C++ code is open-sourced for the benefit of the community. \sjw{It is supposed to assist tasks like surgeon-centered AR, reduction of error, decision making, or safety boundaries for autonomous surgery.}\par

	\sjw{We plan to implement BDIS on GPU end for faster performance. Theoretically, BDIS should be faster than ELAS on GPU.} Future work may also focus on the adaption of the proposed Bayesian strategy to the DNN-based methods. \sjw{It is also interesting to investigate the fusion of BDIS and DNN-based methods because the prior-free CPU-based BDIS and the DNN-based GPU-based methods are complementary.} When DNN is carried out on GPU and dependent on the training data set, CPU-based prior-free BDIS can assist in data fusion or cross-validation. \par

	\section*{Acknowledgment}
	Toyota Research Institute provided funds to support this work. Funding for M. Ghaffari was in part
	provided by NSF Award No. 2118818. \sjw{We would also like to thank Mr. Adam Smidt from University of British Columbia for his helpful suggestion.}

	
	\balance
	{\small
		\bibliographystyle{IEEEtranN}
		\bibliography{bib/strings-abrv,bib/ieee-abrv,reference}
	}
	
\end{document}